\documentclass[11pt]{article}

\usepackage[final]{acl}

\usepackage{times}
\usepackage{latexsym,eucal}

\usepackage[T1]{fontenc}
\usepackage[utf8]{inputenc}

\usepackage{microtype}

\IfFileExists{inconsolata.sty}{\usepackage{inconsolata}}{}

\usepackage{graphicx}
\usepackage{xcolor}
\usepackage{listings}

\usepackage{hyperref}
\usepackage{url}
\IfFileExists{algorithm.sty}{\usepackage{algorithm}}{}
\IfFileExists{algorithmicx.sty}{\usepackage{algorithmicx}}{}
\IfFileExists{algpseudocode.sty}{\usepackage{algpseudocode}}{}
\usepackage{booktabs}
\usepackage{multirow}
\usepackage{wrapfig}
\usepackage{float}
\usepackage{amsmath}
\usepackage{amssymb}
\usepackage{cleveref}

\usepackage{xcolor}

\title{ForeDreamer: A Self-Evolving Dual-Agent Memory Architecture for Future Event Prediction}

\author{
\begin{tabular}{c}
Linhao Zhong$^{1,2}$\thanks{Equal Contribution.} \quad Zongze Du$^{1}$\footnotemark[1] \quad Linyu Wu$^{3}$ \quad Yu Bo$^{1}$ \quad Hourong Li$^{1}$ \\ Chenchen Jing$^{1,4}$ \quad Hao Chen$^{1}$  \quad Yuling Xi$^{1}$\thanks{Corresponding Authors.}  \quad Chunhua Shen$^{1,2,4}$\footnotemark[2] \\[0.2cm]
\textnormal{$^{1}$Zhejiang University, State Lab of CAD \& CG} \quad
\textnormal{$^{2}$Ant Group} \\
\textnormal{$^{3}$National University of Singapore} \quad
\textnormal{$^{4}$Zhejiang University of Technology}
\end{tabular}
}

\begin{document}
\maketitle
\begin{abstract}
% Future event prediction over open-web evidence requires agents to identify relevant search results, filter noisy and redundant observations, and produce calibrated predictions from incomplete and uncertain signals. Existing retrieval and agent-memory mechanisms are often insufficient for this setting, as they either pass retrieved information directly to the model or focus on relatively simple memory functions such as storing and reusing prior information. We propose ForeDreamer, a self-evolving dual-agent framework for managing factual memory over open-web evidence. ForeDreamer distinguishes factual memory, a question-specific evidence interface for the current forecast, from experiential memory, persistent agent memory accumulated across forecasting episodes. A main agent searches for relevant evidence and produces forecasts, while a memory-processing subagent converts search results into factual memory using MemGuides and executable MemTools. ForeDreamer further evolves experiential memory along two tracks: textual forecasting experience for improving forecasting decisions and procedural evidence-processing experience for improving factual-memory construction. Experiments on Prophet Arena and FutureX demonstrate the effectiveness of ForeDreamer for open-web future event prediction.
Open-web future event prediction requires agents to distill reliable signals from noisy, redundant, and incomplete evidence. 
% Existing retrieval and agent-memory mechanisms directly feed retrieved information to agents or rely on simple memory functions such as storing and reusing prior information, leaving them insufficient for open-web forecasting.
Existing retrieval/memory mechanisms directly feed retrieved information to agents or rely on simple memory functions such as storing and reusing prior information for prediction, leaving them insufficient for open-web forecasting.
We propose to transform raw web evidence into structured memory before prediction, enabling agents to reason over distilled, question-specific evidence rather than noisy retrieval results.
This paper presents ForeDreamer, a self-evolving dual-agent framework for managing memory over open-web evidence. 
ForeDreamer separates factual memory, a question-specific evidence state for the current forecast, from experiential memory, persistent agent experience accumulated across forecasting episodes. 
It uses a main agent for search and prediction, and a memory-processing subagent to convert search results into factual memory with dedicated tools. 
% It uses a main agent for search and prediction, and a memory-processing subagent to convert search results into factual memory with MemGuides and executable MemTools. 
ForeDreamer further evolves experiential memory through two tracks, improving both forecasting decisions and factual-memory construction. 
Experiments on Prophet Arena and FutureX demonstrate the effectiveness of ForeDreamer.
\textbf{Project page:} \url{https://zhongzero.github.io/ForeDreamer}
\end{abstract}

\section{Introduction}

Large language model (LLM)~\citep{vaswani2017attention,brown2020language,openai2024gpt4,yang2025qwen3} agents are increasingly deployed in open-world settings that require tool use, web interaction, and adaptation across tasks. A central requirement in these settings is agent memory~\citep{chhikara2025mem0, kang2025memory}, which maintains task-relevant information that supports an agent's behavior within and across tasks. Future event prediction~\citep{yang2025llm, zeng2025futurex} places particular demands on agent memory because an agent must identify relevant open-web evidence, organize noisy and redundant observations into a coherent evidence basis, and produce a calibrated prediction from incomplete and uncertain signals.

\begin{figure}[t]
    \centering
    \includegraphics[width=\columnwidth]{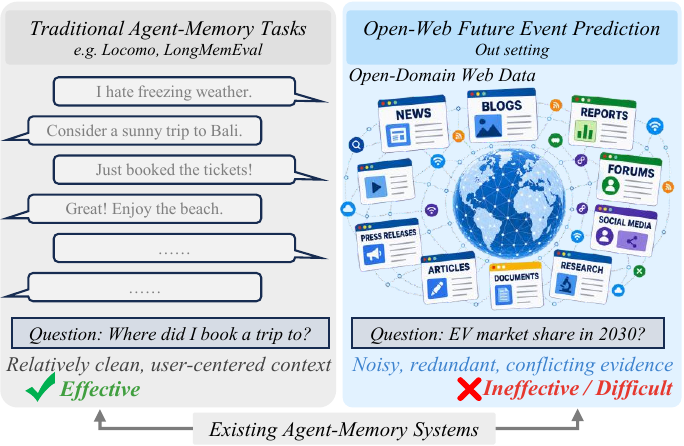}
    \caption{Task mismatch of existing agent-memory systems. Traditional memory tasks often involve relatively clean, user-centered context and can be addressed by storing, retrieving, or reusing prior information. Open-web future event prediction instead requires agents to handle heterogeneous, noisy, and conflicting web data, which existing memory mechanisms are insufficient to process into forecast-ready evidence.}
    \label{fig:memory-task-mismatch}
\end{figure}

Existing retrieval and agent-memory mechanisms~\citep{chhikara2025mem0, kang2025memory, fang2025lightmem} are not sufficient for this setting. In web-augmented forecasting, agents can access search results, but existing approaches often focus on how these results are supplied to the model. A simple long-context strategy largely passes retrieved information directly into the model context, whereas retrieval-augmented generation typically performs lightweight selection or aggregation before generation. Both strategies can remain unreliable when relevant signals are sparse, time-sensitive, and mixed with distracting or conflicting reports. Long-term agent-memory systems, by contrast, are often designed for dialogue personalization, static knowledge reuse, or accumulated user-specific context. Such systems usually emphasize relatively simple memory functions, including the storage, retrieval, or reuse of prior information, and are therefore less suited to the more complex setting of open-web future event prediction. Figure~\ref{fig:memory-task-mismatch} illustrates this task mismatch. Existing memory mechanisms can support relatively clean, user-centered memory tasks~\citep{maharana2024evaluating,wu2024longmemeval}, but they do not directly address the need to process noisy and conflicting open-web evidence.

In this paper, we distinguish between two roles of agent memory in open-ended forecasting. Factual memory denotes a query-conditioned evidence artifact constructed from retrieved web observations for the current forecasting question. It serves as the processed evidence interface for the current forecast, with its scope restricted to the current question. Experiential memory denotes persistent agent memory that accumulates across forecasting episodes and guides future behavior. In open-web forecasting, experiential memory should support both forecasting decisions, including search planning and forecast calibration, and evidence-processing decisions that determine how noisy search artifacts are transformed into factual memory.

We aim to develop an agent-memory framework for open-ended future event prediction by addressing the following two research questions (RQs):
\begin{itemize}
    \item \textbf{RQ1.} \textit{How can an agent make reliable future event predictions from noisy open-web evidence?}
    \item \textbf{RQ2.} \textit{How can feedback from forecasting episodes be used to evolve experiential memory that improves both factual-memory construction and forecasting decisions?}
\end{itemize}

\noindent \textbf{Contribution 1.}
To answer RQ1, we formalize factual-memory management for open-ended future event prediction and propose \textbf{ForeDreamer}, a self-evolving dual-agent framework for managing factual memory over open-web evidence. ForeDreamer separates forecasting from evidence processing. The main agent searches for relevant evidence, integrates processed evidence, and produces the final forecast. A memory-processing subagent receives search results and constructs question-specific factual memory using a MemGuide to specify the evidence-processing workflow and executable MemTools to perform evidence-processing operations. This design exposes factual memory as the question-specific interface between raw evidence and the final forecast.

\noindent \textbf{Contribution 2.}
To answer RQ2, we develop a dual-track evolution mechanism for experiential memory in ForeDreamer. One track updates textual forecasting experience that guides search planning, evidence integration, and forecast calibration. The other track updates procedural evidence-processing experience that guides how search results are converted into factual memory. We further analyze limitations of the initial evolution process, including redundant tool generation and over-concentrated exploration, and introduce two optimizations, compositional tool reuse and diversity-guided exploration scheduling. Experiments on Prophet Arena~\citep{yang2025llm} and FutureX~\citep{zeng2025futurex} demonstrate the effectiveness of ForeDreamer for open-web future event prediction.

\section{ForeDreamer Framework}
\label{sec:method}

\begin{figure*}[t]
    \centering
    \includegraphics[width=\textwidth]{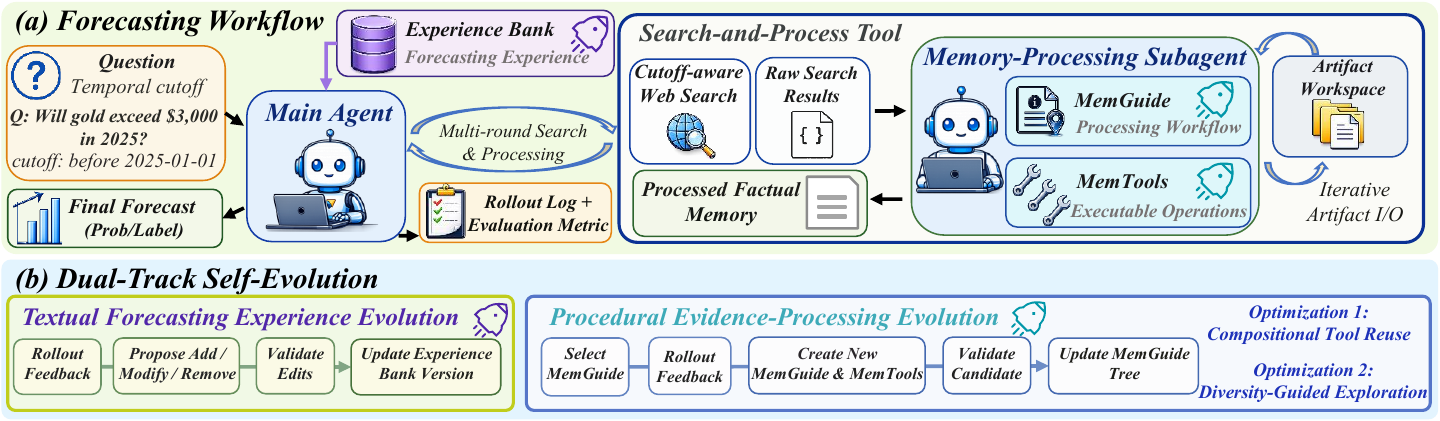}
    \caption{Overview of ForeDreamer. (a) During forecasting, the main agent uses forecasting experience from the Experience Bank, invokes a search-and-process tool to collect temporally valid evidence, and receives processed factual memory from a memory-processing subagent. The subagent follows a MemGuide and uses executable MemTools through an artifact workspace. (b) Rollout feedback drives dual-track self-evolution: textual forecasting experience evolves through validated edits to the Experience Bank, while procedural evidence-processing experience evolves through validated updates to the MemGuide tree and MemTools.}
    \label{fig:foredreamer-overview}
\end{figure*}

ForeDreamer is a self-evolving agent-memory framework for open-web future event prediction. Given a forecasting question, it retrieves temporally valid web evidence, converts raw search results into question-specific factual memory, and predicts from the processed evidence. It separates factual memory, the local evidence interface for the current question, from experiential memory, persistent guidance accumulated across forecasting episodes. Figure~\ref{fig:foredreamer-overview} summarizes the inference workflow and dual-track self-evolution.

\subsection{Task Formulation}
\label{sec:task-formulation}

Each instance consists of a forecasting question $q$ and a temporal cutoff $\tau_q$ specifying the information available at prediction time. The system outputs a forecast $\hat{y}$, either a probability distribution over listed outcomes or a discrete answer, depending on the benchmark. We denote the cutoff-constrained evidence space by $\mathcal{E}_{\leq \tau_q}$. Since open-web results are long, noisy, and only partially relevant, ForeDreamer inserts factual-memory construction between retrieval and forecasting.

\subsection{Architecture Overview}
\label{sec:foredreamer-overview}

As shown in Figure~\ref{fig:foredreamer-overview}, ForeDreamer contains two agents. The main agent receives the question, consults the Experience Bank $\mathcal{B}=\{e_i\}_{i=1}^{|\mathcal{B}|}$, plans search actions, integrates factual memory, and produces the final prediction, where each $e_i$ is a reusable textual forecasting experience. The memory-processing subagent converts raw search-result artifacts into factual memory under a MemGuide $g$ and its associated MemTool set $\mathcal{T}_g$: the MemGuide specifies the evidence-processing workflow, and MemTools provide executable operations. This makes evidence processing explicit and inspectable, rather than leaving the forecasting model to interpret raw search results from a long prompt. Appendix~\ref{sec:appendix-memguide-memtool-interface} provides additional details on the MemGuide--MemTool interface.

\subsection{Forecasting Workflow}
\label{sec:forecasting-workflow}

Panel (a) of Figure~\ref{fig:foredreamer-overview} illustrates the forecasting workflow. Given $(q,\tau_q)$, the main agent conditions on $\mathcal{B}$ and runs a multi-turn search-and-reasoning loop with budget $T$. At turn $t$, it may issue one query $s_t$ through the search-and-process tool, and cutoff-aware web search retrieves
\begin{equation}
    \mathcal{R}_t = \operatorname{Search}(s_t, \mathcal{E}_{\leq \tau_q}).
\end{equation}
Each result $r_{t,k}\in\mathcal{R}_t$ is normalized and written to an artifact workspace. The memory-processing subagent then applies the selected MemGuide and MemTools to produce factual memory,
\begin{equation}
    m_{t,k} = P_{g,\mathcal{T}_g}(r_{t,k}, q),
\end{equation}
where $P_{g,\mathcal{T}_g}$ denotes guide-conditioned processing with executable tools. The subagent may iteratively read and write artifacts, and return $\mathcal{M}_t=\{m_{t,k}: r_{t,k}\in\mathcal{R}_t\}$ to the main agent.

Let $K$ be the number of executed search-and-processing turns, with accumulated factual memory $\mathcal{M}_{1:K}=\bigcup_{t=1}^{K}\mathcal{M}_t$. On the final allowed turn, the main agent answers without calling the search tool:
\begin{equation}
    \hat{y}, \rho = A(q, \mathcal{B}, \mathcal{M}_{1:K}),
\end{equation}
where $A$ is the main agent and $\rho$ is the rationale. ForeDreamer records the rollout, including interactions, artifacts, ground truth, and metrics, as feedback for self-evolution. Additional details on search-and-process decomposition, factual-memory representation, and controlled tool execution are provided in Appendix~\ref{sec:appendix-search-process-decomposition}, Appendix~\ref{sec:appendix-factual-memory-artifacts}, and Appendix~\ref{sec:appendix-tool-execution-boundary}, respectively.

\subsection{Dual-Track Self-Evolution Interfaces}
\label{sec:memory-interfaces}

Panel (b) of Figure~\ref{fig:foredreamer-overview} shows how rollout feedback updates experiential memory through two tracks. The textual track applies validated add, modify, or remove operations to the Experience Bank, improving guidance for search planning, evidence integration, and calibration. The procedural track critiques rollout behavior, derives design requirements, generates candidate MemGuides and MemTools, and admits validated candidates into the guide tree. The next section details these algorithms and the two procedural optimizations: compositional tool reuse and diversity-guided exploration.

\section{Dual-Track Experience Evolution}
\label{sec:evolution-optimization}

ForeDreamer evolves experiential memory on a small evolution-and-validation pool $\mathcal{D}_{\mathrm{evo}}$, whose rollouts provide feedback for memory updates and serve as the admission gate for candidates. Evolution has two tracks: textual forecasting experience for search planning, evidence integration, and calibration, and procedural evidence-processing experience for converting search results into factual memory. This section describes both tracks, with procedural evolution summarized in Figure~\ref{fig:guide-evolution}.

\subsection{Textual Forecasting Experience Evolution}
\label{sec:dual-track-evolution}
\label{sec:experience-evolution}

Textual forecasting experience is stored as a compact experience bank $\mathcal{B}^{(v)}=\{e_i\}_{i=1}^{n_v}$ at version $v$. In each textual-evolution attempt, ForeDreamer selects the currently best validated MemGuide, samples an example from $\mathcal{D}_{\mathrm{evo}}$, runs a rollout with the current bank, and asks an editor to propose $o\in\{\textsc{Add},\textsc{Modify},\textsc{Remove}\}$, yielding $\widetilde{\mathcal{B}} = o(\mathcal{B}^{(v)})$. \textsc{Add} appends an entry, \textsc{Modify} revises one, and \textsc{Remove} deletes an ineffective entry. The candidate bank is evaluated on $\mathcal{D}_{\mathrm{evo}}$ under the selected MemGuide and accepted only when validation improves:
\begin{equation}
    \mathcal{B}^{(v+1)} =
    \begin{cases}
        \widetilde{\mathcal{B}}, & \widetilde{\mathcal{B}}\succ \mathcal{B}^{(v)},\\
        \mathcal{B}^{(v)}, & \text{otherwise},
    \end{cases}
\end{equation}
where $\widetilde{\mathcal{B}}\succ \mathcal{B}^{(v)}$ means better validation score.

\subsection{Procedural Evidence-Processing Evolution}
\label{sec:procedural-evolution}

\begin{figure*}[t]
    \centering
    \includegraphics[width=\textwidth]{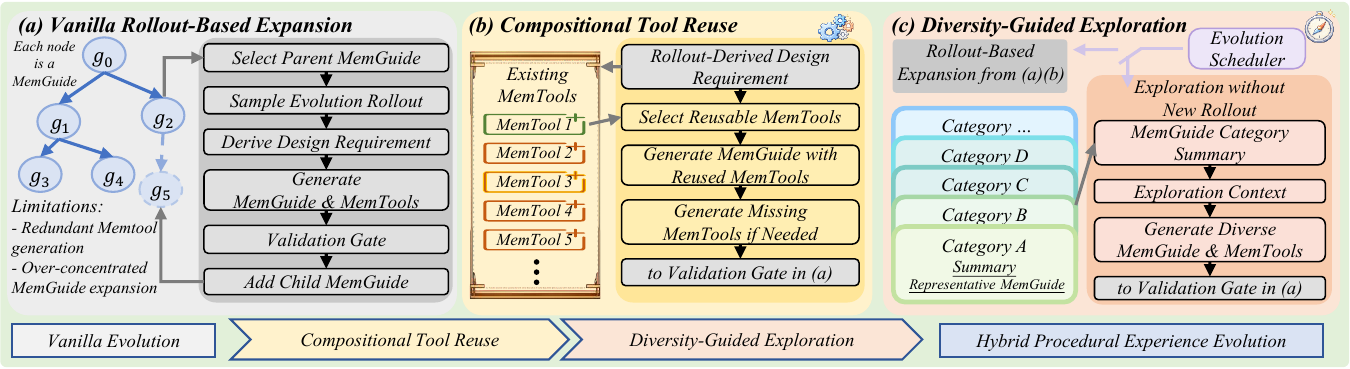}
    \caption{Evolution of procedural evidence-processing experience. (a) Vanilla rollout-based expansion grows a MemGuide tree by selecting a parent MemGuide, sampling an evolution rollout, deriving a design requirement, generating a candidate MemGuide and MemTools, and admitting the candidate through a validation gate. (b) Compositional tool reuse optimizes the candidate-generation step by selecting compatible existing MemTools before generating the MemGuide and any missing MemTools. (c) Diversity-guided exploration uses an evolution scheduler to retain the rollout-based path while adding an exploration path that generates category-diverse candidates from MemGuide-category summaries and representative MemGuides without requiring a new rollout result.}
    \label{fig:guide-evolution}
\end{figure*}

Procedural evidence-processing experience consists of MemGuides, which specify subagent workflows, and MemTools, which provide executable evidence-processing operations. Unlike textual experience, procedural experience must evolve both instructions and tools, and generated assets are accepted only after interface checks and validation. Figure~\ref{fig:guide-evolution} shows vanilla rollout-based expansion and two optimizations.

\subsubsection{Vanilla Rollout-Based Expansion}
\label{sec:vanilla-guide-evolution}

The vanilla mechanism represents procedural experience as a MemGuide tree $\mathcal{G}=(\mathcal{V},\mathcal{E})$, where each node $g_i\in\mathcal{V}$ is a MemGuide and each edge records provenance. In one expansion attempt, ForeDreamer selects a parent MemGuide, samples an example from $\mathcal{D}_{\mathrm{evo}}$, runs a rollout, and uses a critic to derive a design requirement $d$ when improvement is needed. With validation results, parent selection follows validation-rank Zipf sampling: if $g_i$ has rank $r_i$ among $N$ valid nodes, it is selected with probability
\begin{equation}
    p(g_i) = \frac{1 / r_i}{\sum_{j=1}^{N} 1 / r_j}.
\end{equation}
This favors stronger MemGuides while preserving nonzero probability for lower-ranked nodes. ForeDreamer then generates $(\widetilde{g},\widetilde{\mathcal{T}})$ from $d$. If $\operatorname{Valid}(\widetilde{g};\mathcal{D}_{\mathrm{evo}})=1$, the candidate is admitted as a child and the tree is updated as
\begin{equation}
    \mathcal{V}\leftarrow \mathcal{V}\cup\{\widetilde{g}\},\quad
    \mathcal{E}\leftarrow \mathcal{E}\cup\{(g_i,\widetilde{g})\}.
\end{equation}
Validity requires successful validation runs on all examples in $\mathcal{D}_{\mathrm{evo}}$.

The tree provides provenance but can cause redundant MemTool generation and over-concentration around locally successful guide families. ForeDreamer addresses these limitations through compositional tool reuse and diversity-guided exploration.

\subsubsection{Compositional Tool Reuse}
\label{sec:tool-reuse}

\begin{figure}[t]
    \centering
    \includegraphics[width=\columnwidth]{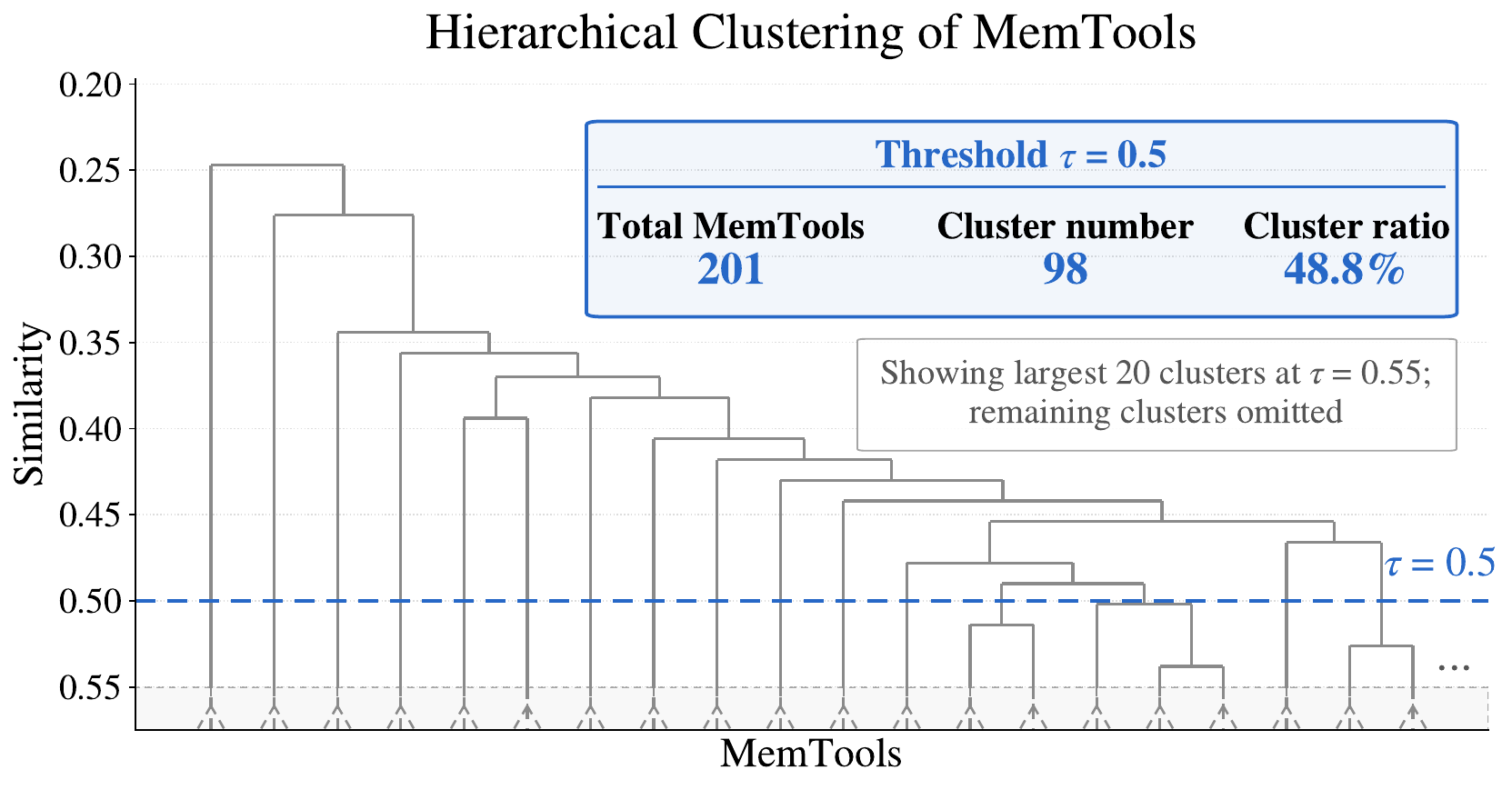}
    \caption{Hierarchical clustering of MemTools generated by vanilla procedural evolution on FutureX, using token-Jaccard source-code similarity. The reduced cluster count indicates that many generated MemTools share substantial token-level overlap, suggesting that vanilla evolution spends many attempts on near-duplicate operations.}
    \label{fig:tool-similarity-token-jaccard}
\end{figure}

\paragraph{Observation.}
Vanilla rollout-based procedural evolution often regenerates MemTools with overlapping logic. In the vanilla FutureX setting, we cluster generated MemTools by pairwise source-code similarity. As shown in Figure~\ref{fig:tool-similarity-token-jaccard}, at token-Jaccard threshold $\tau=0.5$, 201 generated MemTools collapse into 98 clusters, yielding a cluster ratio of 48.8\%. This suggests recurring implementation patterns and wasted evolution budget on near-duplicate operations. Additional similarity views are provided in Appendix~\ref{sec:appendix-tool-similarity-analysis}.

\paragraph{Method.}
ForeDreamer therefore introduces compositional tool reuse, illustrated in Figure~\ref{fig:guide-evolution}(b). Given design requirement $d$, ForeDreamer summarizes existing valid MemTools and selects a compatible subset from the active set, $\mathcal{T}_{\mathrm{reuse}} = R(d,\mathcal{T}_{\mathrm{active}})$ with $\mathcal{T}_{\mathrm{reuse}}\subseteq \mathcal{T}_{\mathrm{active}}$. The candidate MemGuide is newly generated, but it reuses selected tools; new MemTools $\mathcal{T}_{\mathrm{new}}$ are generated only for uncovered operations, yielding $\widetilde{\mathcal{T}}=\mathcal{T}_{\mathrm{reuse}}\cup\mathcal{T}_{\mathrm{new}}$. The candidate returns to the same validation gate, reducing duplicate tool creation while retaining the ability to introduce new tools when needed.

\subsubsection{Diversity-Guided Exploration}
\label{sec:diversity-guided-exploration}

\begin{figure}[t]
    \centering
    \includegraphics[width=\columnwidth]{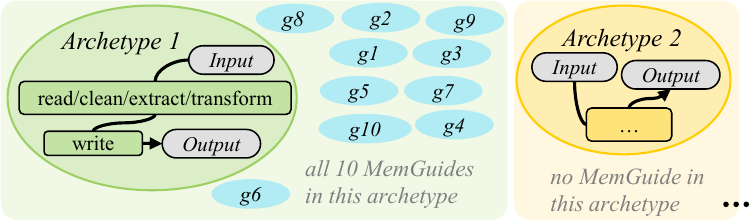}
    \caption{Pipeline archetype distribution for vanilla rollout-based expansion on FutureX. All 10 evolved MemGuides fall into the same tool-orchestrated processing archetype, indicating limited exploration of alternative evidence-processing pipelines.}
    \label{fig:exploration-observation}
\end{figure}

\paragraph{Observation.}
The vanilla MemGuide tree can also constrain exploration. Since each expansion conditions on a parent guide and a local rollout-derived requirement, successful branches may attract nearby variants while other strategies receive fewer trials. On 10 FutureX MemGuides evolved with vanilla expansion, all fall into the same tool-orchestrated pipeline archetype (Figure~\ref{fig:exploration-observation}), indicating refinement of one guide family rather than exploration of alternative evidence-processing pipelines.

\paragraph{Method.}
Figure~\ref{fig:guide-evolution}(c) illustrates diversity-guided exploration. ForeDreamer augments rollout-based expansion with an exploration path that does not require a new rollout result, and a scheduler alternates between exploration and rollout-based expansion. For exploration, ForeDreamer summarizes current MemGuide categories, identifies representative MemGuides, and generates candidates that are not merely local variants of existing branches. Accepted candidates are stored in the same tree, with provenance pointing to the representative guides used as context, and pass through the same validation gate as rollout-derived candidates.

\noindent Together, compositional tool reuse and diversity-guided exploration form a hybrid procedural state with MemGuide-tree provenance, validated reusable MemTools, MemGuide-category summaries, representative MemGuides, and validation records. This preserves tree interpretability while moving procedural evolution beyond local parent-child expansion. Additional details on MemGuide and MemTool validation, procedural summary records, and validation-based candidate selection are provided in Appendix~\ref{sec:appendix-memguide-memtool-validation}, Appendix~\ref{sec:appendix-procedural-summaries}, and Appendix~\ref{sec:appendix-validation-selection}, respectively.

\section{Experiments}
\label{sec:experiments}

\begin{table*}[t]
\centering
\small
\setlength{\tabcolsep}{4pt}
\resizebox{\textwidth}{!}{
\begin{tabular}{llccccccccc}
\toprule
Model & Method & Climate/Weather & Companies & Economics & Entertainment & Mentions & Other & Politics & Sports & Avg. ($\downarrow$) \\
\midrule
\multirow{9}{*}{Qwen3.5-Flash}
& Full Text & 0.3342 & 0.2657 & 0.0720 & 0.2005 & 0.2106 & 0.1306 & 0.2214 & 0.2122 & 0.2059 \\
& RAG & 0.2511 & 0.2319 & 0.0752 & 0.1739 & 0.2178 & 0.1381 & 0.2094 & 0.2030 & 0.1876 \\
& HippoRAG 2 & 0.2798 & 0.2553 & 0.0524 & 0.1318 & 0.2020 & 0.1423 & 0.2069 & 0.1945 & 0.1831 \\
& Mem0 & 0.2792 & 0.2902 & 0.0695 & 0.1936 & 0.2308 & 0.1778 & 0.2423 & 0.2045 & 0.2110 \\
& MemoryOS & 0.2385 & 0.2348 & 0.0542 & 0.1155 & 0.1917 & 0.1676 & 0.2216 & 0.1925 & 0.1771 \\
& A-MEM & 0.2306 & 0.2591 & 0.0602 & 0.1516 & 0.2218 & 0.1542 & 0.2122 & 0.1996 & 0.1862 \\
& LightMem & 0.2562 & 0.2382 & 0.0709 & 0.1253 & 0.1783 & 0.1455 & 0.1971 & 0.1971 & 0.1761 \\
& LangMem & 0.2611 & 0.2516 & 0.0658 & 0.1500 & 0.2190 & 0.1427 & 0.2017 & 0.2024 & 0.1868 \\
& \textbf{ForeDreamer} & \textbf{0.1770} & \textbf{0.2162} & \textbf{0.0414} & \textbf{0.0962} & \textbf{0.1573} & \textbf{0.1274} & \textbf{0.1742} & \textbf{0.1869} & \textbf{0.1471} \\
\midrule
\multirow{9}{*}{GPT-5.4-Nano}
& Full Text & 0.2275 & 0.3719 & 0.0631 & 0.1424 & 0.2878 & 0.1410 & 0.2257 & 0.2075 & 0.2084 \\
& RAG & 0.1856 & 0.3698 & 0.0596 & 0.1377 & 0.2768 & 0.1449 & 0.2216 & 0.2014 & 0.1997 \\
& HippoRAG 2 & 0.2338 & 0.4052 & 0.1216 & 0.1376 & 0.2806 & 0.1515 & 0.2339 & 0.2031 & 0.2209 \\
& Mem0 & 0.2404 & 0.3941 & 0.0582 & 0.2138 & 0.3021 & 0.2009 & 0.2582 & \textbf{0.1782} & 0.2307 \\
& MemoryOS & 0.2039 & 0.3578 & 0.1191 & 0.1455 & 0.2810 & 0.1551 & 0.2026 & 0.1934 & 0.2073 \\
& A-MEM & 0.1980 & 0.4155 & 0.1248 & 0.1782 & 0.2817 & 0.1291 & 0.2273 & 0.1939 & 0.2186 \\
& LightMem & \textbf{0.1584} & 0.4226 & 0.1172 & 0.1609 & 0.2799 & 0.1439 & 0.2117 & 0.2041 & 0.2123 \\
& LangMem & 0.1968 & 0.3654 & 0.0754 & 0.1805 & 0.2844 & 0.1656 & 0.2410 & 0.2076 & 0.2146 \\
& \textbf{ForeDreamer} & 0.1811 & \textbf{0.3456} & \textbf{0.0506} & \textbf{0.1117} & \textbf{0.2704} & \textbf{0.1272} & \textbf{0.1927} & 0.1919 & \textbf{0.1839} \\
\bottomrule
\end{tabular}
}
\caption{Main comparison on Prophet Arena. The table reports Brier score for each category and the average score; lower values indicate better calibrated forecasts.}
\label{tab:main-results-prophet-arena}
\end{table*}

\begin{table}[t]
\centering
\small
\begin{tabular}{llc}
\toprule
Model & Method & Accuracy ($\uparrow$) \\
\midrule
\multirow{9}{*}{Qwen3.5-Flash}
& Full Text & 0.3298 \\
& RAG & 0.3382 \\
& HippoRAG 2 & 0.3285 \\
& Mem0 & 0.3264 \\
& MemoryOS & 0.3269 \\
& A-MEM & 0.3495 \\
& LightMem & 0.2885 \\
& LangMem & 0.3333 \\
& \textbf{ForeDreamer} & \textbf{0.4108} \\
\midrule
\multirow{9}{*}{GPT-5.4-Nano}
& Full Text & 0.2766 \\
& RAG & 0.3221 \\
& HippoRAG 2 & 0.3173 \\
& Mem0 & 0.3567 \\
& MemoryOS & 0.3262 \\
& A-MEM & 0.3480 \\
& LightMem & 0.2788 \\
& LangMem & 0.3333 \\
& \textbf{ForeDreamer} & \textbf{0.3883} \\
\bottomrule
\end{tabular}
\caption{Main comparison on FutureX. The table reports prediction accuracy; higher values are better.}
\label{tab:main-results-futurex}
\end{table}

\subsection{Experimental Setup}
We evaluate ForeDreamer on open-ended future event prediction using Qwen3.5-Flash~\citep{qwen35blog} and GPT-5.4-Nano~\citep{openai2026gpt54} as backbone models. Evaluations are conducted on Prophet Arena~\citep{yang2025llm} and FutureX~\citep{zeng2025futurex}, with temporally constrained web search used for evidence collection. Prophet Arena is evaluated with Brier score, while FutureX is evaluated with accuracy. Baselines include the Full Text setting, a generic RAG baseline, and previously proposed agent-memory systems, including HippoRAG 2~\citep{gutierrez2025rag}, Mem0~\citep{chhikara2025mem0}, MemoryOS~\citep{kang2025memory}, A-MEM~\citep{xu2026mem}, LightMem~\citep{fang2025lightmem}, and LangMem~\citep{langchain2025langmem}. More implementation details and benchmark descriptions are provided in Appendix~\ref{sec:more-implementation-details}, and the prompt templates used in our implementation are provided in Appendix~\ref{sec:key-prompt-templates}.

\subsection{Evaluation Results}

\paragraph{Main performance comparison.}
Table~\ref{tab:main-results-prophet-arena} and Table~\ref{tab:main-results-futurex} present the main benchmark comparison on Prophet Arena and FutureX. On Prophet Arena, ForeDreamer achieves the best average Brier performance among the compared methods for both backbone models, and improves over Full Text across event categories. On FutureX, ForeDreamer also obtains the strongest accuracy among the evaluated methods. These results indicate that evolving task-level forecasting experience together with procedural evidence-processing experience is more effective than directly using retrieved context or applying existing memory baselines to the retrieved evidence. Additional main benchmark results are provided in Appendix~\ref{sec:appendix-additional-main-results}.

\begin{table}[t]
\centering
\small
\resizebox{\columnwidth}{!}{%
\begin{tabular}{lcc}
\toprule
Method & Prophet Arena Avg. ($\downarrow$) & FutureX ($\uparrow$) \\
\midrule
Full Text & 0.2059 & 0.3298 \\
w/o evolving MemGuide\&MemTool & 0.1663 & 0.3892 \\
w/o evolving Experience Bank & 0.1769 & 0.3351 \\
\textbf{ForeDreamer} & \textbf{0.1471} & \textbf{0.4108} \\
\bottomrule
\end{tabular}%
}
\caption{Dual-track experience ablation on Qwen3.5-Flash. Prophet Arena reports average Brier score, while FutureX reports accuracy.}
\label{tab:ablation-guide-experience}
\end{table}

\paragraph{Ablation of dual-track experience.}
Table~\ref{tab:ablation-guide-experience} evaluates the contribution of the two evolution tracks in ForeDreamer. Removing either textual forecasting experience evolution or procedural evidence-processing evolution degrades performance. The full system performs best on both benchmarks, showing that the Experience Bank and the evolved MemGuide--MemTool procedures provide complementary benefits for future event prediction. Additional dual-track experience ablations are provided in Appendix~\ref{sec:appendix-additional-dual-track-ablations}.

\begin{table}[t]
\centering
\small
\resizebox{\columnwidth}{!}{%
\begin{tabular}{lcc}
\toprule
Method & Prophet Arena Avg. ($\downarrow$) & FutureX ($\uparrow$) \\
\midrule
Full Text & 0.2059 & 0.3298 \\
w/o Both Optimizations & 0.1592 & 0.3850 \\
w/o Compositional Tool Reuse & 0.1541 & 0.4032 \\
w/o Diversity-Guided Exploration & 0.1554 & 0.3564 \\
\textbf{ForeDreamer} & \textbf{0.1471} & \textbf{0.4108} \\
\bottomrule
\end{tabular}%
}
\caption{Procedural-evolution optimization ablation on Qwen3.5-Flash. The variants remove Compositional Tool Reuse, Diversity-Guided Exploration, or both; Prophet Arena reports average Brier score, and FutureX reports accuracy.}
\label{tab:ablation-optimization-guide}
\end{table}

\paragraph{Effect of procedural-evolution optimizations.}
Table~\ref{tab:ablation-optimization-guide} isolates the two optimizations used in procedural evidence-processing evolution: Compositional Tool Reuse and Diversity-Guided Exploration. The full design outperforms variants that remove one or both optimizations, indicating that both optimizations contribute to the final performance. Additional procedural-evolution optimization ablations are provided in Appendix~\ref{sec:appendix-additional-procedural-optimization-ablations}.

\begin{table*}[t]
\centering
\small
\setlength{\tabcolsep}{6pt}
\resizebox{\textwidth}{!}{%
\begin{tabular}{lcccc}
\toprule
Setting & Prophet Arena Full Text ($\downarrow$) & Prophet Arena ForeDreamer ($\downarrow$) & FutureX Full Text ($\uparrow$) & FutureX ForeDreamer ($\uparrow$) \\
\midrule
Base Setting & 0.2059 & \textbf{0.1471} & 0.3298 & \textbf{0.4108} \\
Max Interaction Turns = 3 & 0.1871 & \textbf{0.1570} & 0.3894 & \textbf{0.4423} \\
Max Interaction Turns = 4 & 0.1908 & \textbf{0.1629} & 0.3942 & \textbf{0.4279} \\
Firecrawl Search & 0.1995 & \textbf{0.1514} & 0.4394 & \textbf{0.4596} \\
Top-2 Retrieved Results & 0.1831 & \textbf{0.1514} & 0.3544 & \textbf{0.3606} \\
Top-6 Retrieved Results & 0.1845 & \textbf{0.1426} & 0.3398 & \textbf{0.4183} \\
Top-8 Retrieved Results & 0.1811 & \textbf{0.1430} & 0.3503 & \textbf{0.3671} \\
60K Search Context & 0.1833 & \textbf{0.1504} & 0.3930 & \textbf{0.4010} \\
90K Search Context & 0.1855 & \textbf{0.1504} & 0.3831 & \textbf{0.3990} \\
\bottomrule
\end{tabular}%
}
\caption{Robustness across search settings on Qwen3.5-Flash. Each row changes one search or interaction configuration from the base setup and reports the corresponding comparison; Prophet Arena reports average Brier score, and FutureX reports accuracy.}
\label{tab:extend-to-other-settings}
\end{table*}

\paragraph{Robustness across search settings.}
Table~\ref{tab:extend-to-other-settings} evaluates ForeDreamer under alternative search and interaction settings. ForeDreamer consistently improves across these settings on both benchmarks. The results suggest that the observed gains are not tied to a single search provider, retrieval budget, or interaction configuration. Per-category search-setting results are provided in Appendix~\ref{sec:appendix-additional-search-settings}.

\begin{table}[t]
\centering
\small
\resizebox{\columnwidth}{!}{
\begin{tabular}{lccc}
\toprule
Model & No Information & Full Text & ForeDreamer \\
\midrule
\multicolumn{4}{c}{Prophet Arena Avg. ($\downarrow$)} \\
\midrule
Qwen3.5-Flash & 0.2545 & 0.2059 & \textbf{0.1471} \\
GPT-5.4-Nano & 0.2492 & 0.2084 & \textbf{0.1839} \\
\midrule
\multicolumn{4}{c}{FutureX ($\uparrow$)} \\
\midrule
Qwen3.5-Flash & 0.3077 & 0.3298 & \textbf{0.4108} \\
GPT-5.4-Nano & 0.2260 & 0.2766 & \textbf{0.3883} \\
\bottomrule
\end{tabular}
}
\caption{Check for potential data leakage through no-information prompting. The table compares forecasting without web search against settings that use retrieved evidence.}
\label{tab:llm-noleak-clarification}
\end{table}

\paragraph{Check for potential data leakage.}
Table~\ref{tab:llm-noleak-clarification} addresses the concern that the backbone models may have encountered similar forecasting data during pretraining. The No Information setting removes web search and asks the model to forecast from the question alone, while the Full Text setting and ForeDreamer use temporally valid retrieved evidence. The performance gap between No Information and the search-based settings indicates that external web evidence is needed for strong forecasting performance, supporting that the gains are not explained by the backbone model alone. Appendix~\ref{sec:appendix-additional-main-results} further reports results with GPT-5-Nano~\citep{openai2025gpt5}, whose training-data cutoff precedes the evaluated benchmark periods, and Appendix~\ref{sec:appendix-additional-clarifications} provides additional no-information checks.

\begin{figure}[t]
    \centering
    \includegraphics[width=\columnwidth]{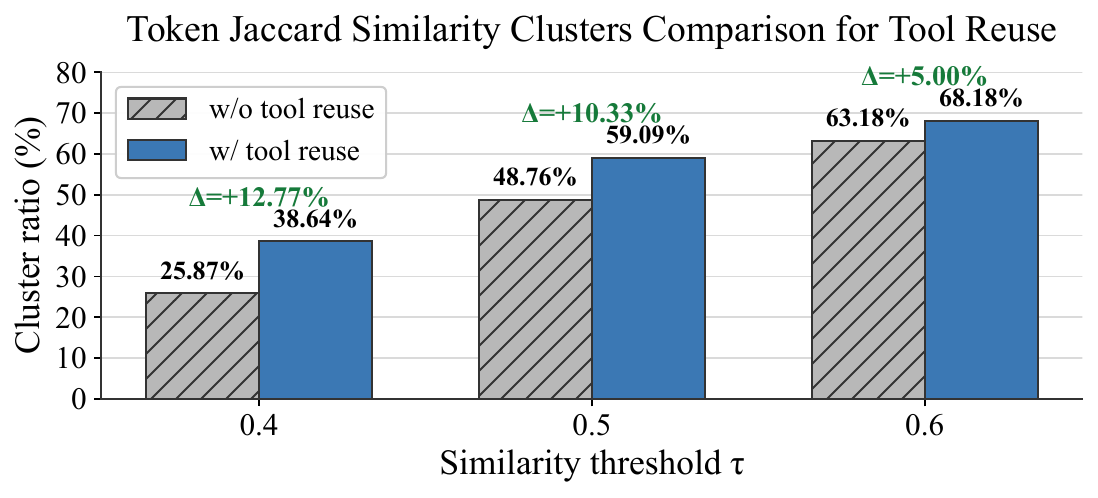}
    \caption{MemTool similarity analysis on FutureX using token-Jaccard source-code similarity. Compositional Tool Reuse produces a less redundant tool set under the same clustering criterion.}
    \label{fig:tool-similarity-token-jaccard-comparison}
\end{figure}

\paragraph{Analysis of Compositional Tool Reuse.}
Beyond aggregate benchmark scores, we analyze how the evolved procedural memory changes under the proposed optimizations. Section~\ref{sec:tool-reuse} reports the token-Jaccard clustering observation that motivates Compositional Tool Reuse. Figure~\ref{fig:tool-similarity-token-jaccard-comparison} compares the generated MemTools on FutureX under the same token-Jaccard setting. The comparison shows that Compositional Tool Reuse produces a less redundant MemTool set, supporting its role in reducing repeated tool construction and improving the efficiency of procedural evolution. Appendix~\ref{sec:appendix-tool-similarity-analysis} provides the metric definitions, additional normalized-sequence and AST-based clustering views, and the tool-reuse cluster-ratio table.

\begin{figure}[t]
    \centering
    \includegraphics[width=\columnwidth]{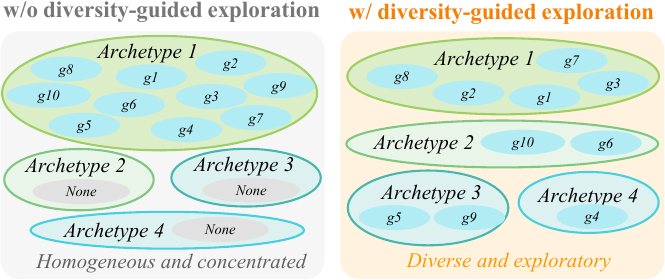}
    \caption{MemGuide pipeline-archetype analysis on FutureX. Diversity-Guided Exploration encourages evolved guides to cover more pipeline archetypes, while vanilla rollout-based expansion remains concentrated in one archetype.}
    \label{fig:exploration-comparison-concise}
\end{figure}

\paragraph{Analysis of Diversity-Guided Exploration.}
Figure~\ref{fig:exploration-comparison-concise} analyzes the effect of Diversity-Guided Exploration on evolved MemGuide pipelines. The comparison shows that the proposed exploration strategy encourages MemGuides to cover more pipeline archetypes rather than repeatedly refining a single guide family. Appendix~\ref{sec:appendix-guide-archetype-analysis} provides detailed archetype definitions and guide assignments.

\section{Related Work}

\subsection{Agent Memory}
Long-horizon agents require memory beyond prompt-local context, especially when preferences, temporal facts, and reusable task experience must persist across sessions~\citep{zhang2025survey, hu2025memory, du2026memory, wu2025human, lin2026survey}. Early memory-augmented language systems largely relied on retrieval-augmented generation and dense or non-parametric memory~\citep{lewis2020retrieval,guu2020retrieval,borgeaud2022improving,packer2023memgpt}, while recent agent-memory work develops more structured storage, updating, and retrieval policies. HippoRAG and HippoRAG~2 organize external knowledge through graph-based associative retrieval for long-term factual and relational recall~\citep{gutierrez2024hipporag,gutierrez2025rag}; Mem0, MemoryOS, A-MEM, and LightMem explore persistent profiles, hierarchical memory management, dynamic memory graphs, and efficient consolidation~\citep{chhikara2025mem0,kang2025memory,xu2026mem,fang2025lightmem}. LangMem represents a complementary engineering direction that integrates semantic, episodic, and procedural memories into agent frameworks~\citep{langchain2025langmem}.

\subsection{Agent Self-Evolving}
% Self-evolving agents~\citep{yang2026ttcs, yue2026dr, he2026active, hou2026learn, weng2026group, guo2026evoconfig} aim to improve their behavior after deployment through reflection, experience reuse, skill discovery, or even modification of their own procedures. Prior work has studied reflection over failed trajectories, iterative output refinement, tool-use learning, and open-ended skill acquisition~\citep{shinn2023reflexion,madaan2023self,schick2023toolformer,wang2023voyager}. In memory-centric agents, this evolution can happen at several levels: evolving the memory operation itself, evolving meta-memory for evidence utilization, evolving the memory architecture, or evaluating continuous test-time memory adaptation. MemSkill learns and revises reusable memory skills~\citep{zhang2026memskill}; MetaMem optimizes a self-reflective meta-memory for knowledge use~\citep{xin2026metamem}; MemEvolve performs meta-evolution over agent memory systems~\citep{zhang2025memevolve}; and Evo-Memory benchmarks streaming test-time learning with self-evolving memory~\citep{wei2025evomemory}. Related work on self-improving coding agents shows that agents can also improve by editing their own implementation rather than only updating natural-language memory~\citep{robeyns2025self}.
Self-evolving agents~\citep{yang2026ttcs, yue2026dr, he2026active, hou2026learn, weng2026group, guo2026evoconfig} improve after deployment through reflection, experience reuse, skill discovery, or procedural modification. Prior work studies failed-trajectory reflection, iterative refinement, tool-use learning, and open-ended skill acquisition~\citep{shinn2023reflexion,madaan2023self,schick2023toolformer,wang2023voyager}. For memory-centric agents, evolution can target reusable memory skills~\citep{zhang2026memskill}, self-reflective meta-memory~\citep{xin2026metamem}, memory-system architectures~\citep{zhang2025memevolve}, or streaming test-time adaptation~\citep{wei2025evomemory}. Self-improving coding agents further show that agents can revise their own implementation rather than only their natural-language memory~\citep{robeyns2025self}.

\subsection{Future Prediction}
% Future prediction is an increasingly important evaluation setting for LLM agents because it tests information gathering, temporal reasoning, uncertainty calibration, and decision-making under unresolved outcomes. Earlier work studies whether language models can approach human forecasting performance and introduces agentic event-forecasting benchmarks such as MIRAI and ForecastBench~\citep{halawi2024approaching,ye2024mirai,karger2024forecastbench}. In contrast to static QA-style evaluation, live forecasting benchmarks reduce contamination by resolving questions only after model predictions are made. Prophet Arena continuously collects real-market forecasting tasks and evaluates probabilistic predictive intelligence under a multi-horizon protocol~\citep{yang2025llm}. FutureX similarly frames future prediction as a dynamic live benchmark for LLM agents, emphasizing real-time updates, tool use, temporal validity, and robustness to misleading web evidence~\citep{zeng2025futurex}. 
% These benchmarks complement traditional long-term memory evaluations such as LoCoMo and LongMemEval~\citep{maharana2024evaluating,wu2024longmemeval} by testing whether stored experience and retrieved evidence can support calibrated forecasts about events that have not yet resolved.
Future prediction evaluates LLM agents on information gathering, temporal reasoning, calibration, and decisions about unresolved events. Earlier work examines whether language models can approach human forecasting and introduces agentic forecasting benchmarks such as MIRAI and ForecastBench~\citep{halawi2024approaching,ye2024mirai,karger2024forecastbench}. Live benchmarks reduce contamination by resolving questions after predictions are made: Prophet Arena evaluates probabilistic predictive intelligence on real-market forecasting tasks~\citep{yang2025llm}, while FutureX studies dynamic future prediction with tool use, temporal validity, and robustness to misleading web evidence~\citep{zeng2025futurex}. 

\section{Conclusion}
We presented ForeDreamer, a self-evolving dual-agent framework for open-web future event prediction. ForeDreamer separates question-specific factual memory from experiential memory accumulated across forecasting episodes, allowing a main forecasting agent to reason over processed evidence while a memory-processing subagent constructs this evidence interface with MemGuides and executable MemTools. It further improves procedural evolution through compositional tool reuse and diversity-guided exploration. Experiments on Prophet Arena and FutureX show that open-web forecasting benefits from treating memory not only as stored information, but also as evolving experience for evidence processing and prediction.

\section*{Limitations}
ForeDreamer is designed for open-web future event prediction, where agents must process noisy search results into forecast-ready factual memory. Evaluating the framework on traditional agent-memory benchmarks is therefore outside the scope of this work, as these benchmarks mainly test whether an agent can store, retrieve, or reuse clean interaction histories. Such benchmarks emphasize different memory functions from the evidence-processing and forecasting setting studied in this paper.
Another limitation is that resolved forecasting feedback is scarce in temporally valid future-event prediction, so ForeDreamer evolves and validates its memory using a small feedback pool. This setting may raise concerns about validation-set overfitting. However, the evolved memories are evaluated on held-out forecasting questions that are not used for evolution or validation, and the observed improvements indicate that ForeDreamer can extract reusable forecasting and evidence-processing patterns from a small number of resolved examples rather than merely fitting individual validation questions. 
% Finally, our empirical comparison focuses on memory-oriented baselines and on analyzing the contribution of self-evolution within ForeDreamer. These comparisons are aligned with the paper's goal of studying how agent memory can evolve for open-web forecasting, but they do not exhaust the broader space of forecasting-agent designs. Future work should compare ForeDreamer with stronger forecasting-specific pipelines and agents.

\section*{Ethical Considerations}
This work focuses on open-web future event prediction using public benchmark datasets. It does not involve collecting private user data, conducting human-subject experiments, or annotating sensitive personal information. All experiments are conducted for research evaluation under temporally constrained forecasting settings. We do not identify additional ethical concerns specific to this work.

% \section*{Acknowledgments}

\bibliography{main,custom}

\clearpage
\appendix
\renewcommand\thesection{\Alph{section}}
\renewcommand\thefigure{S\arabic{figure}}
\renewcommand\thetable{S\arabic{table}}
\renewcommand\theequation{S\arabic{equation}}
\makeatletter
\@ifundefined{thealgorithm}{}{\renewcommand\thealgorithm{S\arabic{algorithm}}}
\makeatother

\renewcommand\theHfigure{S\arabic{figure}}
\renewcommand\theHtable{S\arabic{table}}
\renewcommand\theHequation{S\arabic{equation}}
\makeatletter
\@ifundefined{theHalgorithm}{}{\renewcommand\theHalgorithm{S\arabic{algorithm}}}
\makeatother

\setcounter{figure}{0}
\setcounter{table}{0}
\setcounter{equation}{0}
\makeatletter
\@ifundefined{c@algorithm}{}{\setcounter{algorithm}{0}}
\makeatother

% \section{LLM Usage}

% In this section, we clarify the role of large language models (LLMs) in preparing this work. The model was used exclusively for language polishing, such as refining grammar, style, and readability, without contributing to the research design, analysis, or conclusions.

\section*{Appendix}

\section{Appendix Overview}

This appendix provides additional methodological details, implementation details, experimental results, and analyses to supplement the main paper. It is organized as follows:

\begin{itemize}
    % \item \textbf{Appendix~\ref{sec:more-probability-estimation-formulas}: More Probability Estimation Formulas}
    % \begin{itemize}
    % \end{itemize}

    \item \textbf{Appendix~\ref{sec:more-methodological-details}: More Methodological Details}
    \begin{itemize}
        \item Appendix~\ref{sec:appendix-search-process-decomposition}: Search-and-Process Decomposition
        \item Appendix~\ref{sec:appendix-factual-memory-artifacts}: Factual-Memory Representation
        \item Appendix~\ref{sec:appendix-memguide-memtool-interface}: MemGuide and MemTool Interface
        \item Appendix~\ref{sec:appendix-memguide-memtool-validation}: MemGuide and MemTool Validation
        \item Appendix~\ref{sec:appendix-procedural-summaries}: Procedural Summary Records
        \item Appendix~\ref{sec:appendix-validation-selection}: Validation Records and Candidate Selection
        \item Appendix~\ref{sec:appendix-tool-execution-boundary}: Controlled Tool Execution
    \end{itemize}

    \item \textbf{Appendix~\ref{sec:more-implementation-details}: More Implementation Details}
    \begin{itemize}
        \item Appendix~\ref{sec:appendix-evaluation-benchmarks}: Detailed Descriptions of Evaluation Benchmarks
        \item Appendix~\ref{sec:appendix-baseline-methods}: Detailed Descriptions of Baseline Methods
        \item Appendix~\ref{sec:appendix-dataset-splits-evolution}: Dataset Splits and Evolution Protocol
        \item Appendix~\ref{sec:appendix-inference-evaluation}: Inference and Evaluation Setup
    \end{itemize}

    \item \textbf{Appendix~\ref{sec:more-experimental-results}: More Experimental Results}
    \begin{itemize}
        \item Appendix~\ref{sec:appendix-additional-main-results}: Additional Results on Main Benchmarks
        \item Appendix~\ref{sec:appendix-additional-dual-track-ablations}: Additional Dual-Track Experience Ablations
        \item Appendix~\ref{sec:appendix-additional-procedural-optimization-ablations}: Additional Procedural-Evolution Optimization Ablations
        \item Appendix~\ref{sec:appendix-additional-search-settings}: Additional Results with Different Search Settings
        \item Appendix~\ref{sec:appendix-additional-clarifications}: Additional Data-Leakage Checks
    \end{itemize}

    \item \textbf{Appendix~\ref{sec:more-analyses}: More Analyses}
    \begin{itemize}
        \item Appendix~\ref{sec:appendix-tool-similarity-analysis}: Tool Similarity Analysis
        \item Appendix~\ref{sec:appendix-guide-archetype-analysis}: MemGuide Pipeline Archetype Analysis
    \end{itemize}

    \item \textbf{Appendix~\ref{sec:key-prompt-templates}: Key Prompt Templates}
    \begin{itemize}
        \item Appendix~\ref{sec:appendix-inference-prompts}: Inference Prompts
        \item Appendix~\ref{sec:appendix-evolution-prompts}: Evolution Prompts
    \end{itemize}
\end{itemize}

\section{More Methodological Details}
\label{sec:more-methodological-details}

This section provides additional methodological details for ForeDreamer. We describe how search and evidence processing are decomposed, how factual memory is represented, how MemGuides and MemTools are specified and validated, how procedural summaries and validation records support evolution, and how tool execution is controlled.

\subsection{Search-and-Process Decomposition}
\label{sec:appendix-search-process-decomposition}

ForeDreamer exposes a single search-and-process interface to the main agent, but internally decomposes factual-memory construction into retrieval and processing stages. The retrieval stage issues the query, applies the task-specific temporal cutoff when supported by the search provider, normalizes the returned evidence, and bounds the amount of evidence passed forward. The processing stage then invokes the memory-processing subagent to transform each retrieved item into concise, task-relevant factual memory. This decomposition keeps the main-agent interface simple while preserving a clear methodological separation between evidence acquisition and evidence transformation. The processed evidence is returned to the main agent as factual support for forecasting, rather than as instructions that can alter the agent's control flow.

\subsection{Factual-Memory Representation}
\label{sec:appendix-factual-memory-artifacts}

ForeDreamer represents factual memory at the granularity of individual retrieved items. Each retrieved item is converted into a structured record that preserves the forecasting question, the search query, the temporal constraint, source metadata, and the retrieved content. The memory-processing subagent transforms this raw record into a plain-text factual-memory output that is consumed by the main agent. This representation makes the evidence path auditable: each piece of processed factual memory can be traced back to the retrieved item from which it was derived. When a MemGuide decomposes evidence processing into multiple stages, intermediate artifacts are kept within the same item-level workspace, so multi-step processing remains localized to the corresponding evidence item.

\subsection{MemGuide and MemTool Interface}
\label{sec:appendix-memguide-memtool-interface}

A MemGuide defines the processing objective and constrains the MemTools that the memory-processing subagent may use during a workflow. It does not directly execute operations; instead, it specifies how the subagent should coordinate available tools to transform raw evidence into factual memory. A MemTool implements a bounded evidence-processing operation with a declared callable interface, such as reading source evidence, extracting salient information, creating intermediate representations, validating processed content, or writing the final factual memory. This separation keeps workflow-level instructions and executable operations distinct while allowing procedural experience to combine LLM-based reasoning with tool-based evidence processing.

\subsection{MemGuide and MemTool Validation}
\label{sec:appendix-memguide-memtool-validation}

Generated procedural assets are checked before they are admitted into active procedural memory. ForeDreamer first verifies that the generated MemGuide has the required workflow fields and that every referenced tool is available either as an existing MemTool or as part of the same candidate. It then validates the generated MemTools for syntactic correctness, interface consistency, and compatibility with the tool-calling schema used by the subagent. Candidate assets are staged and loaded together before publication, so a guide and its tools are accepted only as a mutually consistent unit. This prevents partially valid procedural updates from entering the evolving memory state.

\subsection{Procedural Summary Records}
\label{sec:appendix-procedural-summaries}

ForeDreamer maintains lightweight summaries of the procedural memory state. The MemTool summary describes the capabilities and interfaces of valid tools, and Compositional Tool Reuse uses this summary to select compatible existing tools before generating missing operations for a new design requirement. The MemGuide summary records high-level workflow patterns and representative guides. For Diversity-Guided Exploration, candidate guides are compared with representatives according to workflow intent, stage structure, state transitions, intermediate-artifact usage, tool composition, and responsibility allocation across tools. A candidate that matches an existing workflow pattern is assigned to that category; otherwise, it becomes a new representative. These summaries expose procedural evolution history to the generator in a compact form, without requiring all previous tools, guides, or rollouts to be included in the prompt.

\subsection{Validation Records and Candidate Selection}
\label{sec:appendix-validation-selection}

ForeDreamer stores validation records for evolved MemGuides and Experience Bank versions. A validation record is considered admissible only when all validation examples are completed successfully and the task metric can be computed. Candidate ranking follows the benchmark metric: FutureX candidates are ranked by accuracy, while Prophet Arena candidates are ranked by Brier score. Textual-experience versions are distinguished by the content of the active Experience Bank, which avoids conflating validation results from different banks. During procedural evolution, only validation-qualified MemGuide nodes are used for validation-based candidate ranking.

\subsection{Controlled Tool Execution}
\label{sec:appendix-tool-execution-boundary}

ForeDreamer controls tool execution through two design choices. First, the main agent can access only the public search-and-process interface, while internal retrieval and processing tools remain hidden behind the orchestration layer. Second, during factual-memory construction, the memory-processing subagent may call only the MemTools specified by the active MemGuide. This allowlist ensures that a guide defines not only the intended workflow but also the executable action space available to the subagent. Tool execution is further isolated from external network access and unrelated local state, and each call is subject to a bounded execution budget. These constraints keep evolved tools focused on item-level evidence processing and reduce the risk that procedural memory depends on uncontrolled side effects.

\section{More Implementation Details}
\label{sec:more-implementation-details}

This section provides implementation details for the experimental setup. We describe the evaluation benchmarks, the baseline methods, the dataset splits and evolution protocol, and the inference and evaluation settings used in our experiments.

\subsection{Detailed Descriptions of Evaluation Benchmarks}
\label{sec:appendix-evaluation-benchmarks}
We provide additional details on the two forecasting benchmarks used in our evaluation.
\begin{itemize}
    \item \textbf{Prophet Arena} is a live forecasting benchmark designed to evaluate predictive intelligence on real-world events. The original benchmark continuously collects unresolved events from prediction markets, constructs forecasting contexts, asks models to assign probabilities to possible market outcomes, and evaluates the predictions only after the events resolve. Its evaluation emphasizes both statistical forecast quality and decision value, including Brier score and market-return-style metrics. In our experiments, we use a resolved Prophet Arena snapshot containing 1200 forecasting questions from eight categories: Climate/Weather, Companies, Economics, Entertainment, Mentions, Other, Politics, and Sports. The snapshot times range from June 17, 2025 to November 16, 2025. Because the Sports category contains many examples, our actual evaluation uses a 400-example subset for Sports. We report Brier score as the main metric and mean market return across categories as an auxiliary metric.
    \item \textbf{FutureX} is a live benchmark for future prediction by LLM agents. It is built around a dynamic pipeline that gathers future-oriented questions from web sources, records model predictions before resolution, obtains answers after the resolution date, and scores the prior predictions. The benchmark is designed to reduce data contamination by evaluating events that are genuinely unresolved at prediction time and by refreshing the event pool over time. FutureX covers broad real-world domains such as politics, economics, finance, sports, culture, technology, weather, health, and space, and includes multiple event formats, including single-choice, multi-choice, ranking, and numerical prediction. In our experiments, we use a resolved FutureX evaluation split containing 208 questions with end times from January 15, 2026 to March 10, 2026. Each instance provides the forecasting prompt, event title, end time, difficulty level, and ground-truth answer. We report accuracy after matching the model's final answer to the released ground truth.
\end{itemize}

\subsection{Detailed Descriptions of Baseline Methods}
\label{sec:appendix-baseline-methods}
Below, we describe the baseline methods included in our comparison, excluding the Full Text setting:
\begin{itemize}
    \item \textbf{RAG} is a generic retrieval-augmented generation method that conditions generation on retrieved external evidence rather than relying only on parametric model knowledge.
    \item \textbf{HippoRAG 2} is a non-parametric continual-learning framework for LLMs that augments RAG with graph-based memory retrieval, deeper passage integration, and online LLM use.
    \item \textbf{Mem0} is a scalable long-term memory architecture that dynamically extracts, consolidates, and retrieves salient information from ongoing conversations, with an optional graph-memory variant for relational structure.
    \item \textbf{MemoryOS} organizes agent memory through hierarchical short-term, mid-term, and long-term storage together with storage, updating, retrieval, and generation modules.
    \item \textbf{A-MEM} constructs agent memory as an adaptive, interconnected note network inspired by the Zettelkasten method, using dynamic indexing, linking, and memory evolution when new memories are added.
    \item \textbf{LightMem} is a lightweight memory-augmented generation framework that separates sensory, short-term, and long-term memory stages to reduce memory-management overhead while preserving historical interaction information.
    \item \textbf{LangMem} is a LangChain toolkit that provides primitives and tools for extracting, storing, searching, and updating long-term agent memories across interactions.
\end{itemize}
In our experiments, these baseline methods are applied to process the evidence retrieved by the search module before producing the final forecast.

\subsection{Dataset Splits and Evolution Protocol}
\label{sec:appendix-dataset-splits-evolution}
For Prophet Arena, ForeDreamer is evolved independently for each category. In each category, we randomly sample five examples and use them for both evolution and validation, while the remaining examples are reserved for evaluation. For FutureX, we randomly sample 20 examples from the 208 resolved examples for evolution and validation, and use the remaining examples for evaluation.

Each evolution run is allocated 60 update iterations. The update process alternates between procedural evolution, which revises MemGuides and MemTools, and declarative evolution, which revises the Experience Bank. During procedural evolution, Diversity-Guided Exploration is interleaved with rollout-based expansion. 
After evolution, we rank the evolved configurations by validation performance, retain the top five candidates, evaluate them on the held-out test set, and report the best test performance among these candidates.

\subsection{Inference and Evaluation Setup}
\label{sec:appendix-inference-evaluation}
All reported forecasting runs enforce temporal validity during retrieval: search results are restricted to content published before the question-specific cutoff time. Unless otherwise specified, each forecasting run allows at most two interaction turns. The default retrieval setup uses Tavily search, retrieves up to four results per query, and allocates a 30K-character context budget to each search turn. In Table~\ref{tab:extend-to-other-settings} and Table~\ref{tab:extend-to-other-settings-additional}, ``Max Interaction Turns'' varies the interaction budget, ``Firecrawl Search'' changes the search provider, ``Top-$k$ Retrieved Results'' varies the retrieval breadth per query, and ``$N$K Search Context'' varies the per-turn search-context budget.

\section{More Experimental Results}
\label{sec:more-experimental-results}

This section provides additional experimental results that complement the main evaluation in Section~\ref{sec:experiments}. We include results with an additional backbone model, an auxiliary Prophet Arena market-return metric, per-category ablations, per-category search-setting results, and additional checks for potential data leakage.

\subsection{Additional Results on Main Benchmarks}
\label{sec:appendix-additional-main-results}

Table~\ref{tab:main-results-prophet-arena-additional} and Table~\ref{tab:main-results-futurex-additional} provide results with GPT-5-Nano. ForeDreamer achieves the strongest performance among the compared methods on both Prophet Arena and FutureX, consistent with the main benchmark comparison. Because the GPT-5-Nano training-data cutoff precedes the evaluated benchmark periods, these results also provide an additional model-side check against potential data leakage.

\begin{table*}[t]
\centering
\small
\setlength{\tabcolsep}{4pt}
\resizebox{\textwidth}{!}{
\begin{tabular}{llccccccccc}
\toprule
Model & Method & Climate/Weather & Companies & Economics & Entertainment & Mentions & Other & Politics & Sports & Avg. ($\downarrow$) \\
\midrule
\multirow{9}{*}{GPT-5-Nano}
& Full Text & 0.1994 & 0.2870 & 0.1347 & 0.1989 & 0.2225 & 0.1897 & 0.2537 & 0.2066 & 0.2116 \\
& RAG & 0.2438 & 0.2489 & 0.0988 & 0.1884 & 0.1997 & 0.1759 & 0.2658 & 0.2049 & 0.2033 \\
& HippoRAG 2 & 0.2250 & 0.2617 & 0.0987 & 0.2089 & 0.2437 & 0.1656 & 0.2607 & 0.2024 & 0.2083 \\
& Mem0 & 0.2124 & 0.3497 & 0.1373 & 0.2242 & 0.2024 & 0.2378 & 0.2751 & 0.2025 & 0.2302 \\
& MemoryOS & 0.1366 & 0.2603 & 0.1354 & 0.2021 & 0.2412 & 0.2036 & 0.2598 & 0.2063 & 0.2057 \\
& A-MEM & 0.2189 & 0.2786 & 0.1075 & 0.1935 & 0.2142 & 0.2021 & 0.2688 & 0.2032 & 0.2109 \\
& LightMem & 0.1779 & 0.2842 & 0.0764 & \textbf{0.1760} & 0.1989 & \textbf{0.1029} & 0.2583 & 0.2053 & 0.1850 \\
& LangMem & 0.2753 & 0.2438 & 0.0962 & 0.1906 & 0.1954 & 0.1322 & 0.2610 & 0.2019 & 0.1996 \\
& \textbf{ForeDreamer} & \textbf{0.0946} & \textbf{0.1559} & \textbf{0.0476} & 0.1970 & \textbf{0.1798} & 0.1804 & \textbf{0.2517} & \textbf{0.1974} & \textbf{0.1631} \\
\bottomrule
\end{tabular}
}
\caption{Additional Prophet Arena results for GPT-5-Nano. The table reports Brier score for each category and the average score; lower values indicate better calibrated forecasts.}
\label{tab:main-results-prophet-arena-additional}
\end{table*}

\begin{table}[t]
\centering
\small
\begin{tabular}{llc}
\toprule
Model & Method & Accuracy ($\uparrow$) \\
\midrule
\multirow{9}{*}{GPT-5-Nano}
& Full Text & 0.3723 \\
& RAG & 0.3942 \\
& HippoRAG 2 & 0.3894 \\
& Mem0 & 0.3776 \\
& MemoryOS & 0.3667 \\
& A-MEM & 0.3510 \\
& LightMem & 0.3077 \\
& LangMem & 0.3654 \\
& \textbf{ForeDreamer} & \textbf{0.4149} \\
\bottomrule
\end{tabular}
\caption{Additional FutureX results for GPT-5-Nano. The table reports prediction accuracy; higher values are better.}
\label{tab:main-results-futurex-additional}
\end{table}

Table~\ref{tab:main-results-prophet-arena-average-return} reports Prophet Arena mean market return across categories for Qwen3.5-Flash. ForeDreamer obtains the strongest mean market return among the compared methods, which is consistent with the Brier-score conclusion in the main text.

\begin{table}[t]
\centering
\small
\begin{tabular}{lc}
\toprule
Method & Market Return ($\uparrow$) \\
\midrule
Full Text & 0.5790 \\
RAG & 0.5798 \\
HippoRAG 2 & 0.6038 \\
Mem0 & 0.4763 \\
MemoryOS & 0.4901 \\
A-MEM & 0.5679 \\
LightMem & 0.5505 \\
LangMem & 0.7324 \\
\textbf{ForeDreamer} & \textbf{0.7521} \\
\bottomrule
\end{tabular}
\caption{Additional Prophet Arena average market-return results across categories on Qwen3.5-Flash; higher values are better.}
\label{tab:main-results-prophet-arena-average-return}
\end{table}

\subsection{Additional Dual-Track Experience Ablations}
\label{sec:appendix-additional-dual-track-ablations}

Table~\ref{tab:ablation-guide-experience-prophet-arena-qwen-additional}, Table~\ref{tab:ablation-guide-experience-prophet-arena-additional}, and Table~\ref{tab:ablation-guide-experience-futurex-additional} provide additional dual-track experience ablations. The Prophet Arena tables give per-category Brier scores for Qwen3.5-Flash and GPT-5.4-Nano, while the FutureX table reports GPT-5.4-Nano accuracy. ForeDreamer obtains the best average performance in these comparisons, while removing either the MemGuide--MemTool evolution or the Experience Bank evolution degrades the overall result. These results are consistent with the main ablation and further support the complementarity of the two evolution tracks.

\begin{table*}[t]
\centering
\small
\setlength{\tabcolsep}{4pt}
\resizebox{\textwidth}{!}{
\begin{tabular}{lccccccccc}
\toprule
Method & Climate/Weather & Companies & Economics & Entertainment & Mentions & Other & Politics & Sports & Avg. ($\downarrow$) \\
\midrule
Full Text & 0.3342 & 0.2657 & 0.0720 & 0.2005 & 0.2106 & 0.1306 & 0.2214 & 0.2122 & 0.2059 \\
w/o evolving MemGuide\&MemTool & 0.2328 & 0.2199 & 0.0800 & 0.1125 & 0.1678 & \textbf{0.1274} & 0.1843 & 0.2056 & 0.1663 \\
w/o evolving Experience Bank & 0.2440 & 0.2492 & 0.0666 & 0.1351 & 0.1869 & 0.1306 & 0.2035 & 0.1996 & 0.1769 \\
\textbf{ForeDreamer} & \textbf{0.1770} & \textbf{0.2162} & \textbf{0.0414} & \textbf{0.0962} & \textbf{0.1573} & \textbf{0.1274} & \textbf{0.1742} & \textbf{0.1869} & \textbf{0.1471} \\
\bottomrule
\end{tabular}
}
\caption{Additional dual-track experience ablation on Prophet Arena for Qwen3.5-Flash. The table reports Brier score; lower values are better.}
\label{tab:ablation-guide-experience-prophet-arena-qwen-additional}
\end{table*}

\begin{table*}[t]
\centering
\small
\setlength{\tabcolsep}{4pt}
\resizebox{\textwidth}{!}{
\begin{tabular}{lccccccccc}
\toprule
Method & Climate/Weather & Companies & Economics & Entertainment & Mentions & Other & Politics & Sports & Avg. ($\downarrow$) \\
\midrule
Full Text & 0.2275 & 0.3719 & 0.0631 & 0.1424 & 0.2878 & 0.1410 & 0.2257 & 0.2075 & 0.2084 \\
w/o evolving MemGuide\&MemTool & 0.2275 & 0.3719 & 0.0963 & 0.1349 & 0.2878 & 0.1397 & 0.2222 & 0.2003 & 0.2101 \\
w/o evolving Experience Bank & 0.2251 & \textbf{0.3456} & 0.0547 & 0.1551 & 0.2915 & 0.1546 & 0.2230 & 0.2075 & 0.2071 \\
\textbf{ForeDreamer} & \textbf{0.1811} & \textbf{0.3456} & \textbf{0.0506} & \textbf{0.1117} & \textbf{0.2704} & \textbf{0.1272} & \textbf{0.1927} & \textbf{0.1919} & \textbf{0.1839} \\
\bottomrule
\end{tabular}
}
\caption{Additional dual-track experience ablation on Prophet Arena for GPT-5.4-Nano. The table reports Brier score; lower values are better.}
\label{tab:ablation-guide-experience-prophet-arena-additional}
\end{table*}

\begin{table}[t]
\centering
\small
\begin{tabular}{lc}
\toprule
Method & Accuracy ($\uparrow$) \\
\midrule
Full Text & 0.2766 \\
w/o evolving MemGuide\&MemTool & 0.2766 \\
w/o evolving Experience Bank & 0.3404 \\
\textbf{ForeDreamer} & \textbf{0.3883} \\
\bottomrule
\end{tabular}
\caption{Additional dual-track experience ablation on FutureX for GPT-5.4-Nano. The table reports prediction accuracy; higher values are better.}
\label{tab:ablation-guide-experience-futurex-additional}
\end{table}

\subsection{Additional Procedural-Evolution Optimization Ablations}
\label{sec:appendix-additional-procedural-optimization-ablations}

Table~\ref{tab:ablation-optimization-guide-additional} expands the procedural-evolution optimization ablation on Prophet Arena into per-category results. ForeDreamer achieves the best average Brier score among the compared variants, while removing both optimizations or either individual optimization leads to weaker average performance. This per-category table supports the main conclusion that both optimizations contribute to the final system.

\begin{table*}[t]
\centering
\small
\setlength{\tabcolsep}{4pt}
\resizebox{\textwidth}{!}{
\begin{tabular}{lccccccccc}
\toprule
Method & Climate/Weather & Companies & Economics & Entertainment & Mentions & Other & Politics & Sports & Avg. ($\downarrow$) \\
\midrule
Full Text & 0.3342 & 0.2657 & 0.0720 & 0.2005 & 0.2106 & 0.1306 & 0.2214 & 0.2122 & 0.2059 \\
w/o Both Optimizations & 0.2319 & 0.1948 & 0.0471 & 0.1096 & 0.1917 & \textbf{0.1169} & 0.1881 & 0.1935 & 0.1592 \\
w/o Compositional Tool Reuse & \textbf{0.1531} & 0.2149 & 0.0621 & 0.1038 & 0.1890 & 0.1226 & 0.1965 & 0.1907 & 0.1541 \\
w/o Diversity-Guided Exploration & 0.1747 & \textbf{0.1935} & 0.0602 & 0.1023 & 0.1823 & 0.1439 & 0.1921 & 0.1942 & 0.1554 \\
\textbf{ForeDreamer} & 0.1770 & 0.2162 & \textbf{0.0414} & \textbf{0.0962} & \textbf{0.1573} & 0.1274 & \textbf{0.1742} & \textbf{0.1869} & \textbf{0.1471} \\
\bottomrule
\end{tabular}
}
\caption{Additional procedural-evolution optimization ablation on Prophet Arena for Qwen3.5-Flash. The table reports Brier score; lower values are better.}
\label{tab:ablation-optimization-guide-additional}
\end{table*}

\subsection{Additional Results with Different Search Settings}
\label{sec:appendix-additional-search-settings}

Table~\ref{tab:extend-to-other-settings-additional} reports per-category Prophet Arena results under additional search and interaction settings. ForeDreamer lowers the average Brier score in every setting, indicating that its gains are robust to changes in interaction budget, search provider, retrieval breadth, and search-context budget.

\begin{table*}[t]
\centering
\small
\setlength{\tabcolsep}{4pt}
\resizebox{\textwidth}{!}{
\begin{tabular}{llccccccccc}
\toprule
Setting & Method & Climate/Weather & Companies & Economics & Entertainment & Mentions & Other & Politics & Sports & Avg. ($\downarrow$) \\
\midrule
\multirow{2}{*}{Base Setting}
& Full Text & 0.3342 & 0.2657 & 0.0720 & 0.2005 & 0.2106 & 0.1306 & 0.2214 & 0.2122 & 0.2059 \\
& \textbf{ForeDreamer} & \textbf{0.1770} & \textbf{0.2162} & \textbf{0.0414} & \textbf{0.0962} & \textbf{0.1573} & \textbf{0.1274} & \textbf{0.1742} & \textbf{0.1869} & \textbf{0.1471} \\
\midrule
\multirow{2}{*}{Max Interaction Turns = 3}
& Full Text & 0.2577 & 0.2437 & 0.0955 & 0.1354 & 0.2051 & 0.1639 & 0.2016 & 0.1941 & 0.1871 \\
& \textbf{ForeDreamer} & \textbf{0.2165} & \textbf{0.2227} & \textbf{0.0655} & \textbf{0.0930} & \textbf{0.1741} & \textbf{0.1286} & \textbf{0.1710} & \textbf{0.1844} & \textbf{0.1570} \\
\midrule
\multirow{2}{*}{Max Interaction Turns = 4}
& Full Text & 0.2838 & \textbf{0.2196} & 0.1283 & 0.1830 & 0.1999 & 0.1408 & \textbf{0.1814} & 0.1897 & 0.1908 \\
& \textbf{ForeDreamer} & \textbf{0.2480} & 0.2276 & \textbf{0.0538} & \textbf{0.0992} & \textbf{0.1672} & \textbf{0.1263} & 0.1974 & \textbf{0.1834} & \textbf{0.1629} \\
\midrule
\multirow{2}{*}{Firecrawl Search}
& Full Text & 0.2901 & 0.2377 & 0.1354 & 0.1370 & 0.2056 & 0.1885 & 0.2016 & 0.2001 & 0.1995 \\
& \textbf{ForeDreamer} & \textbf{0.1818} & \textbf{0.2019} & \textbf{0.0515} & \textbf{0.0927} & \textbf{0.1782} & \textbf{0.1098} & \textbf{0.1979} & \textbf{0.1970} & \textbf{0.1514} \\
\midrule
\multirow{2}{*}{Top-2 Retrieved Results}
& Full Text & 0.2254 & 0.2514 & 0.1025 & 0.1173 & 0.2083 & 0.1451 & 0.2171 & \textbf{0.1975} & 0.1831 \\
& \textbf{ForeDreamer} & \textbf{0.2195} & \textbf{0.2183} & \textbf{0.0456} & \textbf{0.0842} & \textbf{0.1505} & \textbf{0.1064} & \textbf{0.1794} & 0.2075 & \textbf{0.1514} \\
\midrule
\multirow{2}{*}{Top-6 Retrieved Results}
& Full Text & 0.2290 & \textbf{0.2158} & 0.1081 & 0.1372 & 0.2202 & 0.1639 & 0.1905 & 0.2114 & 0.1845 \\
& \textbf{ForeDreamer} & \textbf{0.1158} & 0.2286 & \textbf{0.0516} & \textbf{0.0930} & \textbf{0.1580} & \textbf{0.1153} & \textbf{0.1804} & \textbf{0.1980} & \textbf{0.1426} \\
\midrule
\multirow{2}{*}{Top-8 Retrieved Results}
& Full Text & 0.2359 & 0.2300 & 0.0466 & 0.1620 & 0.1770 & 0.1791 & 0.2269 & \textbf{0.1913} & 0.1811 \\
& \textbf{ForeDreamer} & \textbf{0.1847} & \textbf{0.1955} & \textbf{0.0335} & \textbf{0.0944} & \textbf{0.1657} & \textbf{0.0919} & \textbf{0.1783} & 0.2003 & \textbf{0.1430} \\
\midrule
\multirow{2}{*}{60K Search Context}
& Full Text & 0.2417 & \textbf{0.2488} & 0.0641 & 0.1438 & 0.2063 & 0.1618 & 0.1957 & 0.2041 & 0.1833 \\
& \textbf{ForeDreamer} & \textbf{0.1782} & \textbf{0.2488} & \textbf{0.0524} & \textbf{0.1066} & \textbf{0.1481} & \textbf{0.0995} & \textbf{0.1769} & \textbf{0.1923} & \textbf{0.1504} \\
\midrule
\multirow{2}{*}{90K Search Context}
& Full Text & 0.2711 & 0.2312 & 0.0682 & 0.1251 & 0.1983 & 0.1792 & 0.2242 & \textbf{0.1868} & 0.1855 \\
& \textbf{ForeDreamer} & \textbf{0.1908} & \textbf{0.2041} & \textbf{0.0488} & \textbf{0.0904} & \textbf{0.1608} & \textbf{0.1182} & \textbf{0.1869} & 0.2035 & \textbf{0.1504} \\
\bottomrule
\end{tabular}
}
\caption{Per-category results for additional search settings on Prophet Arena with Qwen3.5-Flash (Brier score; lower is better).}
\label{tab:extend-to-other-settings-additional}
\end{table*}

\subsection{Additional Data-Leakage Checks}
\label{sec:appendix-additional-clarifications}

Table~\ref{tab:llm-noleak-clarification-qwen-additional} and Table~\ref{tab:llm-noleak-clarification-gpt54-additional} give per-category Prophet Arena results for the no-information comparison. The No Information setting removes web search, while the other settings use retrieved evidence. The average results show that access to external evidence improves forecasting performance across the evaluated backbones, supporting that the models rely on web evidence rather than only on memorized parametric knowledge.

\begin{table*}[t]
\centering
\small
\setlength{\tabcolsep}{4pt}
\resizebox{\textwidth}{!}{
\begin{tabular}{lccccccccc}
\toprule
Method & Climate/Weather & Companies & Economics & Entertainment & Mentions & Other & Politics & Sports & Avg. ($\downarrow$) \\
\midrule
No Information & 0.2576 & 0.3837 & 0.1225 & 0.2634 & 0.2576 & 0.2390 & 0.2945 & 0.2177 & 0.2545 \\
Full Text & 0.3342 & 0.2657 & 0.0720 & 0.2005 & 0.2106 & 0.1306 & 0.2214 & 0.2122 & 0.2059 \\
\textbf{ForeDreamer} & \textbf{0.1770} & \textbf{0.2162} & \textbf{0.0414} & \textbf{0.0962} & \textbf{0.1573} & \textbf{0.1274} & \textbf{0.1742} & \textbf{0.1869} & \textbf{0.1471} \\
\bottomrule
\end{tabular}
}
\caption{Per-category data-leakage check on Prophet Arena for Qwen3.5-Flash. The table reports Brier score; lower values are better.}
\label{tab:llm-noleak-clarification-qwen-additional}
\end{table*}

\begin{table*}[t]
\centering
\small
\setlength{\tabcolsep}{4pt}
\resizebox{\textwidth}{!}{
\begin{tabular}{lccccccccc}
\toprule
Method & Climate/Weather & Companies & Economics & Entertainment & Mentions & Other & Politics & Sports & Avg. ($\downarrow$) \\
\midrule
No Information & 0.2043 & 0.4272 & 0.1441 & 0.1973 & 0.2911 & 0.2310 & 0.2774 & 0.2210 & 0.2492 \\
Full Text & 0.2275 & 0.3719 & 0.0631 & 0.1424 & 0.2878 & 0.1410 & 0.2257 & 0.2075 & 0.2084 \\
\textbf{ForeDreamer} & \textbf{0.1811} & \textbf{0.3456} & \textbf{0.0506} & \textbf{0.1117} & \textbf{0.2704} & \textbf{0.1272} & \textbf{0.1927} & \textbf{0.1919} & \textbf{0.1839} \\
\bottomrule
\end{tabular}
}
\caption{Per-category data-leakage check on Prophet Arena for GPT-5.4-Nano. The table reports Brier score; lower values are better.}
\label{tab:llm-noleak-clarification-gpt54-additional}
\end{table*}

\section{More Analyses}
\label{sec:more-analyses}

This section provides additional analyses that support the optimization design in ForeDreamer. We include a source-code similarity analysis for MemTools and a pipeline-archetype analysis for evolved MemGuides.

\subsection{Tool Similarity Analysis}
\label{sec:appendix-tool-similarity-analysis}

This appendix details the source-code similarity analysis used to diagnose redundant MemTool generation in the vanilla FutureX setting. We analyze all generated tools in the vanilla configuration without compositional tool reuse or diversity-guided exploration. The analysis contains 201 tools and 20{,}100 pairwise comparisons. For clustering, we use average-linkage agglomerative clustering over pairwise similarities and report the cluster count at similarity threshold $\tau=0.5$.

\paragraph{Similarity metrics.}
We compute three complementary source-code similarities. \textit{Token Jaccard} tokenizes each Python tool with the standard Python tokenizer, removes comments and layout-only tokens, and computes the Jaccard overlap between the two token sets, i.e., $|T_i \cap T_j| / |T_i \cup T_j|$. This is the metric used in the main text because it captures shared implementation vocabulary while being robust to local reordering. \textit{Normalized sequence similarity} uses the same normalized token sequence, but compares two full token sequences with a sequence-matching ratio; it is more sensitive to ordering and near-copy structure. \textit{AST sequence similarity} parses each tool into a Python AST, records the sequence of AST node types visited by \texttt{ast.walk}, and compares these node-type sequences with the same sequence-matching ratio; it focuses on structural similarity rather than exact lexical overlap.

\begin{figure*}[t]
    \centering
    \includegraphics[width=\textwidth]{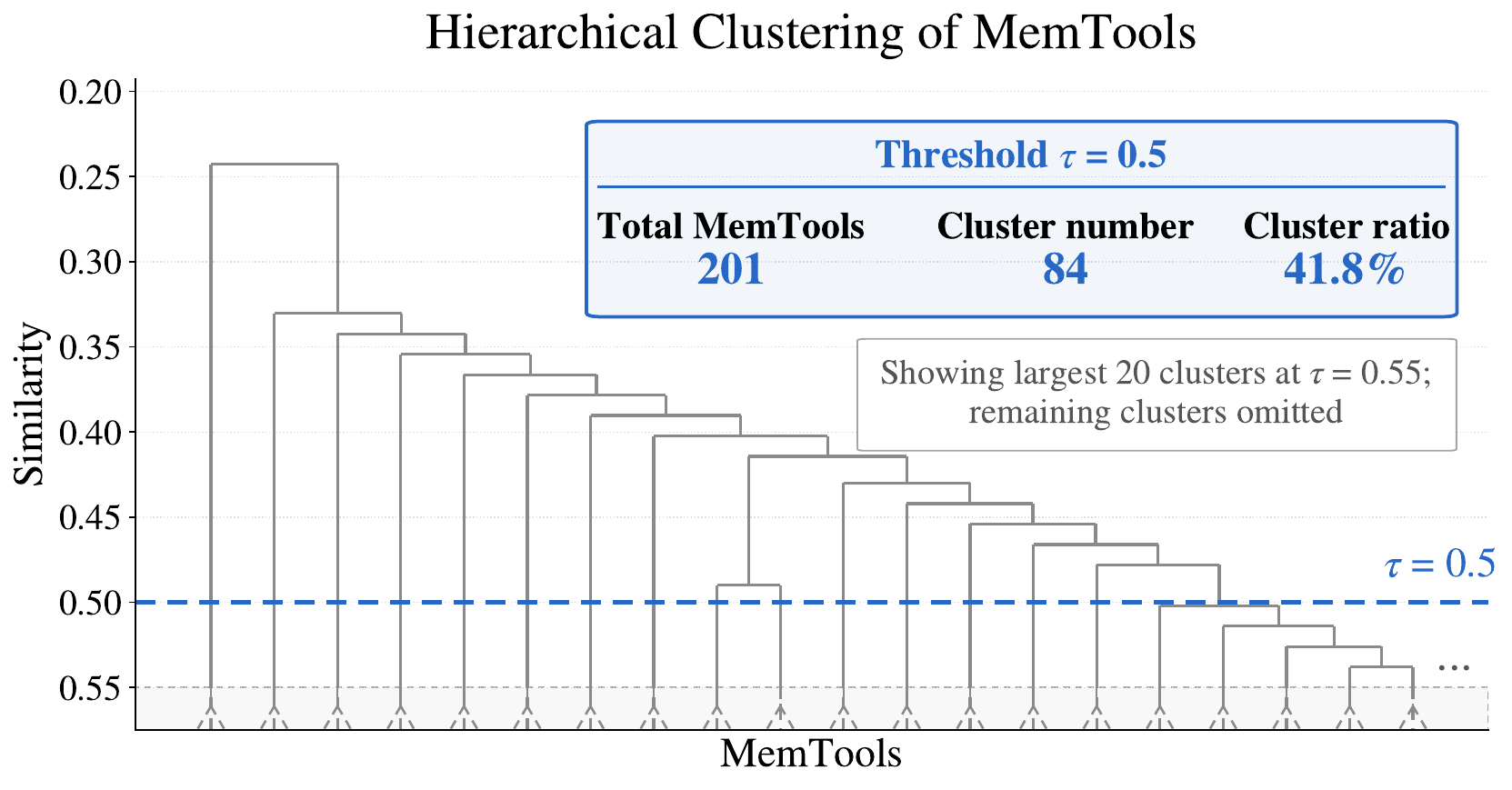}
    \caption{Hierarchical clustering of vanilla FutureX MemTools using normalized token-sequence similarity. The result shows substantial sequence-level overlap among generated tools, complementing the token-Jaccard analysis in the main text.}
    \label{fig:tool-similarity-normalized}
\end{figure*}

\begin{figure*}[t]
    \centering
    \includegraphics[width=\textwidth]{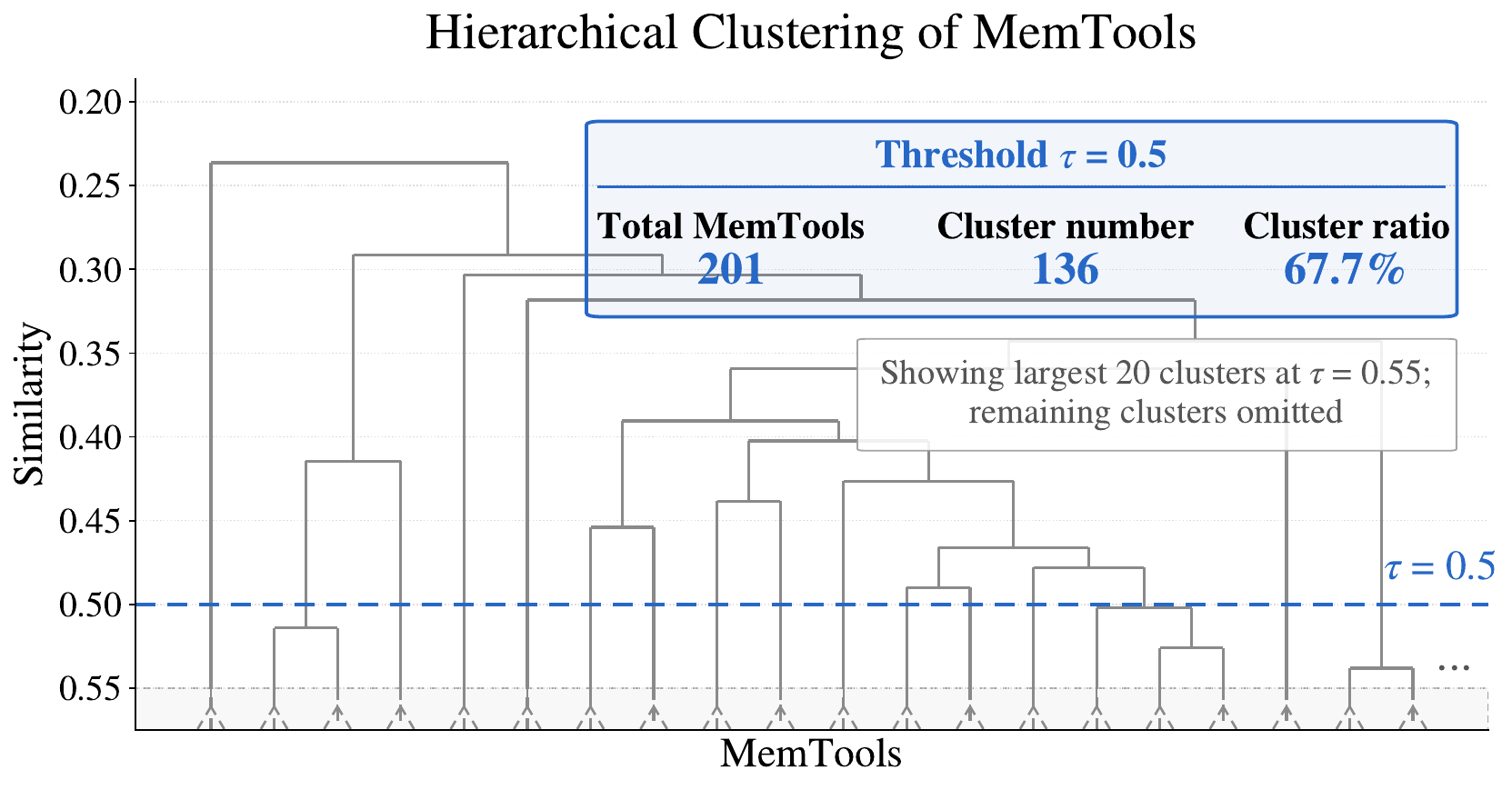}
    \caption{Hierarchical clustering of vanilla FutureX MemTools using AST node-sequence similarity. Even when lexical details are abstracted away, non-singleton clusters remain, indicating repeated structural patterns in generated tools.}
    \label{fig:tool-similarity-ast}
\end{figure*}

\paragraph{Additional clustering results.}
The main text reports token-Jaccard clustering because it offers the most interpretable evidence of repeated tool-level implementation vocabulary. The additional metrics lead to the same qualitative conclusion that vanilla procedural evolution produces many reusable or near-reusable tools. Under normalized sequence similarity at $\tau=0.5$, the 201 tools form 84 clusters, corresponding to a cluster ratio of 41.8\%. Under AST sequence similarity at $\tau=0.5$, the tools form 136 clusters, corresponding to a cluster ratio of 67.7\%. Figure~\ref{fig:tool-similarity-normalized} and Figure~\ref{fig:tool-similarity-ast} show the corresponding clustering views.

\paragraph{Effect of compositional tool reuse.}
Table~\ref{tab:tool-reuse-similarity-analysis} compares the cluster ratios before and after enabling compositional tool reuse on Qwen3.5-Flash. Across both FutureX and Prophet Arena, tool reuse consistently increases the cluster ratio, meaning that the evolved tool set is less aggressively collapsed into similarity clusters. This supports the intended effect of reducing redundant tool generation.

\begin{table*}[t]
\centering
\scriptsize
\resizebox{\textwidth}{!}{%
\begin{tabular}{lllccc}
\toprule
Dataset & Similarity metric & $\tau$ & w/o tool reuse & w/ tool reuse & $\Delta$ \\
\midrule
FutureX & AST sequence & 0.4 & 54.73\% & 68.18\% & \textcolor{green!50!black}{+13.46\%} \\
FutureX & AST sequence & 0.5 & 67.66\% & 79.55\% & \textcolor{green!50!black}{+11.88\%} \\
FutureX & AST sequence & 0.6 & 83.58\% & 84.09\% & \textcolor{green!50!black}{+0.51\%} \\
FutureX & Normalized sequence & 0.4 & 23.88\% & 36.36\% & \textcolor{green!50!black}{+12.48\%} \\
FutureX & Normalized sequence & 0.5 & 41.79\% & 45.45\% & \textcolor{green!50!black}{+3.66\%} \\
FutureX & Normalized sequence & 0.6 & 57.21\% & 63.64\% & \textcolor{green!50!black}{+6.42\%} \\
FutureX & Token Jaccard & 0.4 & 25.87\% & 38.64\% & \textcolor{green!50!black}{+12.77\%} \\
FutureX & Token Jaccard & 0.5 & 48.76\% & 59.09\% & \textcolor{green!50!black}{+10.33\%} \\
FutureX & Token Jaccard & 0.6 & 63.18\% & 68.18\% & \textcolor{green!50!black}{+5.00\%} \\
\midrule
Prophet Arena & AST sequence & 0.4 & 54.73\% & 64.21\% & \textcolor{green!50!black}{+9.48\%} \\
Prophet Arena & AST sequence & 0.5 & 70.14\% & 73.95\% & \textcolor{green!50!black}{+3.81\%} \\
Prophet Arena & AST sequence & 0.6 & 80.52\% & 81.63\% & \textcolor{green!50!black}{+1.11\%} \\
Prophet Arena & Normalized sequence & 0.4 & 20.66\% & 33.26\% & \textcolor{green!50!black}{+12.60\%} \\
Prophet Arena & Normalized sequence & 0.5 & 38.92\% & 49.98\% & \textcolor{green!50!black}{+11.06\%} \\
Prophet Arena & Normalized sequence & 0.6 & 57.67\% & 62.55\% & \textcolor{green!50!black}{+4.88\%} \\
Prophet Arena & Token Jaccard & 0.4 & 22.06\% & 36.37\% & \textcolor{green!50!black}{+14.31\%} \\
Prophet Arena & Token Jaccard & 0.5 & 46.54\% & 55.66\% & \textcolor{green!50!black}{+9.12\%} \\
Prophet Arena & Token Jaccard & 0.6 & 63.98\% & 68.09\% & \textcolor{green!50!black}{+4.11\%} \\
\bottomrule
\end{tabular}%
}
\caption{Effect of compositional tool reuse on MemTool similarity clusters. Values are cluster ratios after converting to percentages; higher ratios indicate that fewer tools are merged into similarity clusters and therefore suggest less redundant tool generation. $\Delta$ is computed as w/ tool reuse minus w/o tool reuse.}
\label{tab:tool-reuse-similarity-analysis}
\end{table*}

\subsection{MemGuide Pipeline Archetype Analysis}
\label{sec:appendix-guide-archetype-analysis}

\begin{figure*}[t]
    \centering
    \includegraphics[width=\textwidth]{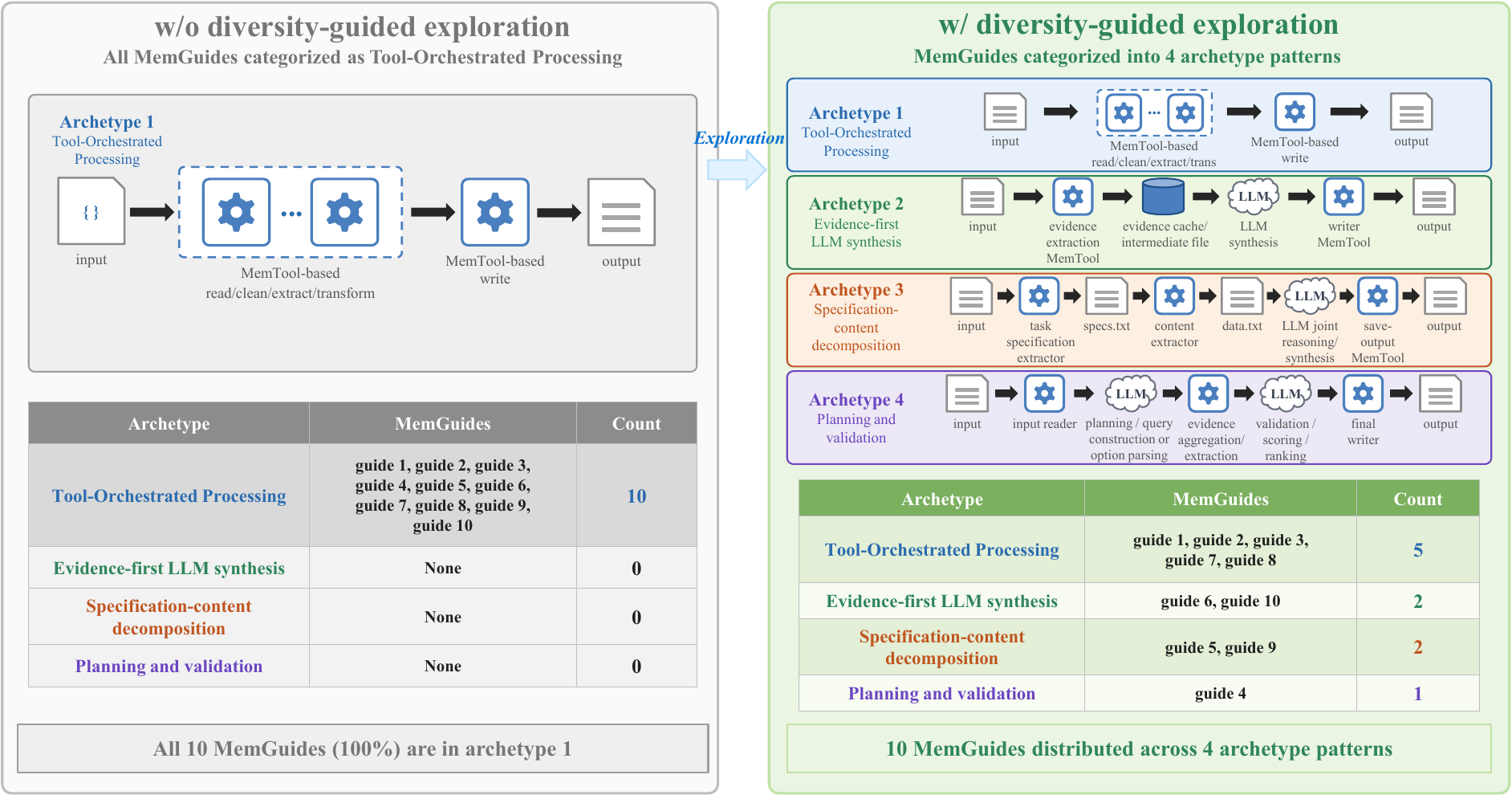}
    \caption{Detailed pipeline-archetype analysis for FutureX MemGuides. Without diversity-guided exploration, all 10 MemGuides fall into Tool-Orchestrated Processing. With diversity-guided exploration, the guides cover four archetype patterns, showing broader procedural exploration.}
    \label{fig:exploration-comparison-detailed}
\end{figure*}

This appendix provides the detailed pipeline-archetype analysis used in Section~\ref{sec:experiments}. We analyze 10 MemGuides evolved on FutureX and categorize each guide by its overall evidence-processing pipeline rather than by the surface wording of the guide. Figure~\ref{fig:exploration-comparison-detailed} shows the resulting archetype assignments before and after enabling diversity-guided exploration.

\paragraph{Pipeline archetypes.}
We identify four recurring pipeline archetypes:
\begin{itemize}
    \item \textbf{Tool-Orchestrated Processing} uses MemTools for the main evidence-processing path. The pipeline reads the input, applies MemTool-based reading, cleaning, extraction, and transformation, and then writes the final output through a MemTool.
    \item \textbf{Evidence-first LLM synthesis} first extracts evidence with a MemTool and stores it in an intermediate cache or file. The LLM then synthesizes the cached evidence, and a writer MemTool produces the final output.
    \item \textbf{Specification-content decomposition} separates task specification processing from content extraction. A specification extractor writes task requirements into a specification file, a content extractor writes evidence content into a data file, and the LLM jointly synthesizes the two sources before a save-output MemTool writes the result.
    \item \textbf{Planning and validation} decomposes the pipeline into explicit planning and checking stages. The guide reads the input, performs planning, query construction, or option parsing, aggregates or extracts evidence, validates, scores, or ranks candidate evidence, and then writes the final output.
\end{itemize}

\paragraph{Archetype distribution.}
Without diversity-guided exploration, all 10 MemGuides are categorized as Tool-Orchestrated Processing. This confirms that vanilla rollout-based expansion concentrates on one pipeline family. After enabling diversity-guided exploration, the 10 MemGuides are distributed across four archetypes: five guides use Tool-Orchestrated Processing, two use Evidence-first LLM synthesis, two use Specification-content decomposition, and one uses Planning and validation. This broader distribution indicates that the exploration path successfully expands procedural evolution beyond local variants of the initially dominant guide pattern.

\clearpage
\onecolumn

\section{Key Prompt Templates}
\label{sec:key-prompt-templates}

\lstdefinestyle{promptstyle}{
    basicstyle=\ttfamily\scriptsize,
    breaklines=true,
    breakatwhitespace=false,
    columns=fullflexible,
    keepspaces=true,
    frame=single,
    framerule=0.4pt,
    rulecolor=\color{gray},
    backgroundcolor=\color{gray!5},
    captionpos=b
}

\subsection{Inference Prompts}
\label{sec:appendix-inference-prompts}

\begin{lstlisting}[style=promptstyle,caption={Main-agent multi-turn search prompt.}]
System message:
You may solve the task over multiple turns and can use the {tool_name} tool when helpful. {cutoff_text} The {tool_name} tool searches the web and returns processed evidence rather than raw results. That processed evidence may still be truncated, so extract only the most relevant information. You may issue at most one {tool_name} tool call in a single assistant turn. Do not batch multiple search requests in one turn. If you need more evidence, use a later turn instead of issuing multiple tool calls at once. Only your final answer should follow the task-specific output format.

Final-turn user message:
This is your final allowed assistant turn. Do not call any tool. Produce the final answer now using only the required task-specific answer format.
\end{lstlisting}

\begin{lstlisting}[style=promptstyle,caption={Experience Bank prompt section.}]
Experience Bank:
These are reusable prior experiences for the main agent. Use them when relevant, but do not let them override direct evidence from the current task.
- [{experience_id}] {experience_text}
\end{lstlisting}

\begin{lstlisting}[style=promptstyle,caption={FutureX forecasting prompt.}]
You are an agent that predicts future events. You may reason over multiple turns and use web_search_and_process to gather supporting evidence when helpful. The web_search_and_process tool returns processed evidence, not raw search results. {search_note} In your final response, provide a detailed rationale section that evaluates and cites the most relevant processed evidence, explains why that evidence is relevant or limited, discusses uncertainties or conflicting signals, and then end with the boxed final answer.

{experience_bank_section}

Event to be predicted:
{question_text}

Final answer requirements:
{answer_requirements}

Response format for this run:
Rationale: <a detailed explanation that cites and evaluates the most relevant evidence, explains its limitations, and justifies the final answer>
\boxed{<final answer>}
Do not put any text after the boxed answer.

Constraint for the final boxed answer line only:
{final_constraints}
\end{lstlisting}

\begin{lstlisting}[style=promptstyle,caption={Prophet Arena forecasting prompt.}]
You are an AI forecasting assistant. You may reason over multiple turns and use web_search_and_process to gather additional evidence when useful. The web_search_and_process tool returns processed evidence, not raw search results. {search_note} In your final rationale, evaluate and cite the most relevant processed evidence instead of giving only a brief generic explanation. Only your final response should be the required JSON object.

{experience_bank_section}

Question:
{question_context}

Market context:
{market_context}

Evidence:
{sources_context}

Final answer requirements:
Return exactly one JSON object with this schema and no markdown fences:
{
  "rationale": "{rationale_placeholder}",
  "probabilities": {
    "{market_name_1}": <probability_value_from_0_to_1>,
    "{market_name_2}": <probability_value_from_0_to_1>
  }
}

Requirements:
- Use exactly the listed outcome names, case-sensitive.
- Do not invent additional outcomes.
- Every listed outcome must appear exactly once.
- Every probability must be a number between 0 and 1.
- {rationale_requirement}
\end{lstlisting}

\begin{lstlisting}[style=promptstyle,caption={Memory-processing subagent prompt.}]
System message:
You are a data-processing subagent. For the current item_dir, use the available MemTool tools to read the required input file, process it, and ensure the required output file is written successfully. Do not invent file contents. Stop once the output file has been created.

User message:
Current item_dir: {item_dir}
Input filename: {input_filename}
Output filename: {output_filename}
Create {output_filename} for this item_dir.
\end{lstlisting}

\subsection{Evolution Prompts}
\label{sec:appendix-evolution-prompts}

\begin{lstlisting}[style=promptstyle,caption={Shared MemGuide/MemTool authoring rules.}]
You are working inside the Memory-Evolving-For-Future-Prediction repository to create or update one `MemGuide`, and optionally create or update one or more `MemTool` files, for the data-process subagent.

Your goal is to define a complete, executable, unambiguous data-processing workflow for a single `item_dir`. The workflow must:
- start from `item_dir/input_filename`
- use one or more MemTool steps to extract, organize, clean, or transform relevant information
- allow a MemTool to read source information and let the LLM directly analyze, compress, rewrite, or generate the answer in the conversation
- optionally create intermediate files under `item_dir`
- optionally let a MemTool directly write the LLM's final answer into a file
- finally write the required result to `item_dir/output_filename`

Exploration and design preferences:
- Prefer workflows with explicit stage boundaries instead of repeatedly collapsing to the shortest possible "one reader tool + LLM + one writer tool" pattern.
- Favor decomposing complex processing into 3 or more bounded stages when appropriate, for example: read, extract, reorganize, save intermediate state, reread intermediate state, compress, validate, and finally write.
- Favor MemTools with clearly different responsibilities rather than repeatedly generating near-duplicate reader or writer tools that differ only in names, field labels, or return wording.
- Favor explicit saving and rereading of intermediate files inside `item_dir` so workflow state transitions are visible, traceable, and reusable.
- When new tools are needed, prefer tools with clearly distinct roles such as: raw-input reader, structured-field extractor, content splitter, intermediate-summary writer, intermediate-summary reader, fragment merger, output validator, or final writer.
- Do not treat small prompt rewrites, slight field-format changes, or renamed variants of existing `read_*` / `write_*` tools as meaningful novelty.

Definitions:
- `MemGuide`: a JSON workflow definition for the data-process subagent. It contains the subagent prompt and the list of MemTools the subagent is allowed to call.
- `MemTool`: a Python tool callable by the data-process subagent. Each MemTool performs one bounded processing step inside the current `item_dir`.
- `item_dir`: the working directory for one data item. The workflow starts from files inside this directory and must also write its final result inside this directory.
- `input_filename`: the main input file for the workflow. The current default example is `raw_data.json`. It is a UTF-8 encoded JSON file. In the current example structure, it contains at least a top-level `item` object, and `item` contains at least:
  - `title`
  - `url`
  - `content`
  - `published_date`
  - `score`
  The most important, most stable, and default-required field is `item.content`, which should be treated as a string. The current example may also contain top-level metadata fields such as:
  - `task_id`
  - `run_label`
  - `dataset_name`
  - `run_dir_name`
  - `task_dir_name`
  - `search_turn`
  - `item_rank`
  - `query`
  - `problem_statement`
  - `task_requirements`
  - `search_before`
  - `content_source`
  - `result_count`
- `output_filename`: the required final output file for the workflow. The current default example is `final_data.txt`. It is a UTF-8 encoded plain-text file, not JSON. Its contents should be the final extracted, transformed, cleaned, or organized text derived from `input_filename`. The workflow is only complete when this file is written under the current `item_dir`.

Follow these requirements strictly:
- Always keep the workflow unambiguous and executable.
- Do not depend on interactive input.
- Do not depend on network access.
- Use ASCII by default in code.
- Do not assume extra files exist outside the sandbox.

Interpret the input and output formats according to these facts:
- the current default input is `raw_data.json`
- the current default output is `final_data.txt`
- `raw_data.json` is structured JSON input
- `final_data.txt` is final plain-text output
- if your design depends on `raw_data.json`, it must at minimum handle `item.content` correctly
- if your design uses additional fields, that usage must remain compatible with the JSON structure described above
- when `raw_data.json` contains `problem_statement` and `task_requirements`, treat them as the upstream task problem and requirement description for the current item
- for FutureX, these fields correspond to the event question plus answer requirements and final boxed-answer constraints
- for Prophet Arena, these fields correspond to the prediction question plus the final JSON output requirements
- a valid workflow may be: a MemTool reads the input, the LLM directly produces the answer in the conversation, and another MemTool writes that answer to `output_filename`

Runtime environment and constraints:
- Every MemTool runs inside a `bwrap` sandbox subprocess.
- The current `item_dir` is readable and writable.
- `/tmp` is a private writable temporary directory inside the sandbox, but it is not part of the business output directory.
- `src/`, the Python runtime, and system directories are read-only.
- Do not depend on writing outside `item_dir`.
- Do not depend on network access.

MemGuide requirements:
- A MemGuide must be a JSON object.
- It must contain at least:
  - `guide_name`
  - `prompt`
  - `tool_names`
- `guide_name` must be non-empty.
- `prompt` must be non-empty.
- `tool_names` must be a non-empty list.
- Every name in `tool_names` must correspond to a MemTool available to the subagent.
- The `prompt` must explicitly require that the subagent:
  - works only for the current `item_dir`
  - uses the provided MemTool set to finish the task
  - does not invent file contents
  - stops once the target output file has been created
- A MemGuide does not perform file I/O itself. It defines how the subagent should use the MemTools to complete the workflow.
- The MemGuide description of the workflow must remain consistent with the input/output formats above. Do not describe `output_filename` as JSON, and do not assume `input_filename` has an arbitrary undefined format.
- A MemGuide may use any of the following patterns, or a combination of them:
  - a tool reads input and another tool directly processes and writes output
  - a tool reads input, the LLM directly reasons or generates the answer in the conversation, and a tool then writes that answer to the output file
  - a tool reads input and writes intermediate files, and later tools or the LLM use those intermediate results to produce the final output
- When the task allows it, prefer multi-stage workflows with explicit intermediate state, such as "read raw input -> extract and write intermediate state -> reread intermediate state and compress/reorganize -> validate or finalize -> write final output".

MemTool requirements:
- Every MemTool must be a Python file corresponding to path format `src/MemTool/tool_<name>.py`
- Every MemTool must define these exports:
  - `TOOL_NAME: str`
  - `TOOL_SPEC: dict[str, Any]`
  - `build_runner_kwargs(arguments, config, runtime_context) -> dict[str, Any]`
  - `run_tool(**kwargs) -> str`
  - `__all__ = ["TOOL_NAME", "TOOL_SPEC", "build_runner_kwargs", "run_tool"]`
- Every MemTool `TOOL_SPEC` must follow OpenAI function-tool style:
  - the top level must be `{"type": "function", "function": {...}}`
  - `function.name` must equal `TOOL_NAME`
  - `function.parameters` must use a JSON Schema object
- Every MemTool `run_tool(...)` must:
  - return a string only
  - raise clear `ValueError` messages for missing input files, invalid formats, or invalid arguments
  - perform business file reads and writes only inside the current `item_dir`
- An individual MemTool does not have to read `item_dir`, `input_filename`, and `output_filename` directly.
- However, the overall guide workflow must be compatible with a runtime context that provides:
  - `item_dir`
  - `input_filename`
  - `output_filename`
- If a MemTool directly reads `input_filename`, it should treat it as a UTF-8 JSON file and should prioritize logic around `item.content`.
- If `problem_statement` and `task_requirements` exist, the MemTool or the LLM may use them to process the current item more accurately.
- If a MemTool directly writes `output_filename`, it should write UTF-8 plain text, not JSON.
- If a MemTool needs specific runtime parameters, read only the fields it actually needs and validate them explicitly.
- A MemTool may implement only one stage of the workflow. It does not need to complete the whole workflow by itself.
- Reader-style MemTools are allowed: they may only read `input_filename` or intermediate files and return the relevant information to the LLM.
- Writer-style MemTools are allowed: they may only take the LLM's current final text and write it to `output_filename`.
- Intermediate-state MemTools are also allowed, for example tools that:
  - write extracted structured data into intermediate files inside `item_dir`
  - reread summaries, fragments, candidate answers, or checklists from intermediate files
  - merge, sort, filter, validate, or rewrite intermediate results
- If you create multiple tools, try to give them meaningfully different roles across stages, data shapes, or state transitions instead of duplicating the same "read original input" or "write final output" function.

Recommended minimal skeleton:

```python
#!/usr/bin/env python3

from pathlib import Path
from typing import Any, Final


TOOL_NAME: Final[str] = "tool_example"
TOOL_SPEC: Final[dict[str, Any]] = {
    "type": "function",
    "function": {
        "name": TOOL_NAME,
        "description": "Describe what this MemTool does.",
        "parameters": {
            "type": "object",
            "properties": {},
            "additionalProperties": False,
        },
    },
}


def build_runner_kwargs(arguments: dict[str, Any], config: Any, runtime_context: Any) -> dict[str, Any]:

    item_dir = str(getattr(runtime_context, "item_dir", "") or "").strip()
    if not item_dir:
        raise ValueError(f"{TOOL_NAME} requires a non-empty runtime item_dir")

    return {
        "item_dir": item_dir,
    }


def run_tool(*, item_dir: str) -> str:
    item_dir_path = Path(item_dir)
    if not item_dir_path.exists():
        raise ValueError(f"{TOOL_NAME} could not find item_dir: {item_dir_path}")

    return f"{TOOL_NAME} completed its step"


__all__ = ["TOOL_NAME", "TOOL_SPEC", "build_runner_kwargs", "run_tool"]
```

If your design needs direct access to the main input or output files, you may also use this style:

```python
def build_runner_kwargs(arguments: dict[str, Any], config: Any, runtime_context: Any) -> dict[str, Any]:
    del arguments
    del config

    item_dir = str(getattr(runtime_context, "item_dir", "") or "").strip()
    input_filename = str(getattr(runtime_context, "input_filename", "") or "").strip()
    output_filename = str(getattr(runtime_context, "output_filename", "") or "").strip()
    if not item_dir:
        raise ValueError(f"{TOOL_NAME} requires a non-empty runtime item_dir")
    if not input_filename:
        raise ValueError(f"{TOOL_NAME} requires a non-empty runtime input_filename")
    if not output_filename:
        raise ValueError(f"{TOOL_NAME} requires a non-empty runtime output_filename")

    return {
        "item_dir": item_dir,
        "input_filename": input_filename,
        "output_filename": output_filename,
    }
```

Special attention:
- `input_filename = raw_data.json`
- `output_filename = final_data.txt`
- in the most basic example, `final_data.txt` may simply equal `raw_data.json["item"]["content"]`
- in richer examples, `raw_data.json` may also include `problem_statement` and `task_requirements`
- your workflow may still perform better extraction, cleaning, compression, summarization, or transformation, as long as the final output remains plain text and is traceable to `input_filename`
\end{lstlisting}

\begin{lstlisting}[style=promptstyle,caption={Textual forecasting experience evolution prompt.}]
You are evaluating whether the main agent's Experience Bank should be updated after one forecasting rollout.

Your job is to inspect:
1. the current Experience Bank (a simplified view that only lists the active experience_id/text pairs),
2. the current best MemGuide validation context,
3. the final run result for the sampled training question,
4. the rollout summary, which includes task metadata, main-agent logs, ground truth, and evaluation feedback.

You must reason about:
- which information in the rollout is directly relevant to the correct answer,
- which parts of the model behavior conflict with the correct answer,
- whether the current information could have been used more effectively to answer the question,
- whether any insight is general and reliable enough to become a reusable experience,
- whether any existing experience is invalid, misleading, redundant, or should be rewritten.

Current Experience Bank:
{{CURRENT_EXPERIENCE_BANK}}

Note: `CURRENT_EXPERIENCE_BANK` only contains the currently active experiences. Each entry only includes `experience_id` and `text`.

Current best guide context:
{{SELECTED_GUIDE_CONTEXT}}

Final run result:
{{RUN_RESULT}}

Rollout summary:
{{ROLLOUT_SUMMARY}}

Note: `ROLLOUT_SUMMARY` only contains the main agent's rollout information. It does not include subagent logs.

Return exactly one JSON object with no markdown fences and no extra text.

Rules:
- Return at most {{MAX_SUGGESTIONS}} suggestions.
- Suggestions must already be ordered from highest priority to lowest priority.
- Use integer priority values where a smaller number means a higher priority.
- `priority = 1` is the highest priority, `priority = 2` is lower, and so on.
- The system will try suggestions in priority order and stop as soon as one suggestion is accepted, so put the most promising suggestion first.
- Valid operations are only `add`, `remove`, and `modify`.
- `remove` and `modify` must reference an existing `target_experience_id`.
- When you use `target_experience_id`, copy it exactly from an `experience_id` field shown in `CURRENT_EXPERIENCE_BANK`.
- Never invent, rename, abbreviate, or paraphrase an existing `target_experience_id`.
- `add` and `modify` must provide a concrete `new_text`.
- Only suggest changes that are genuinely generalizable and likely to improve validation.
- If there are no worthwhile changes, return an empty suggestions list.

Required JSON schema:
{
  "suggestions": [
    {
      "priority": <integer_priority_starting_from_1_where_1_is_highest>,
      "operation": "<add_or_remove_or_modify>",
      "target_experience_id": "<required_for_remove_or_modify_else_empty_string>",
      "new_text": "<required_for_add_or_modify_else_empty_string>",
      "analysis": "<short but concrete reason grounded in the rollout>",
      "generality_assessment": "<why this is general enough or why the old experience is invalid>",
      "expected_benefit": "<how this should improve future validation>"
    }
  ]
}
\end{lstlisting}

\begin{lstlisting}[style=promptstyle,caption={Procedural critic prompt.}]
You are evaluating whether a MemGuide and its available MemTools process retrieved web evidence well enough for downstream forecasting.

Your job is not to write code directly. Your job is to inspect:
1. the current MemGuide,
2. the full source code of the tools available to that guide,
3. a processed rollout summary showing how the guide and tools behaved on one training question.

Decide whether the current guide/tool strategy is already good enough.

If it is good enough:
- return `"should_evolve": false`
- explain why in `"analysis"`
- return an empty string for `"design_requirement"`

If it is not good enough:
- return `"should_evolve": true`
- explain what is insufficient in `"analysis"`
- return a concrete `"design_requirement"` that can be passed into a separate MemGuide/MemTool generator
- the design requirement must describe how to improve the current guide/tool strategy while staying compatible with the same overall MemTool/MemGuide framework
- the design requirement may reuse existing tools and may also request new tools when necessary
- do not return guide JSON or tool code directly

Evaluation criteria:
- whether the processed output captures the most relevant information from the retrieved source
- whether the guide/tool workflow loses important evidence, keeps too much noise, or formats the result poorly
- whether the workflow follows the question-specific requirements well enough
- whether the workflow is robust and general rather than overfitting to one exact example
- whether the current tools are sufficient or need to be extended

Current MemGuide JSON:
{{GUIDE_JSON}}

Current MemTool source code:
{{TOOL_SOURCE_CODE}}

Processed rollout summary:
{{ROLLOUT_SUMMARY}}

Return exactly one JSON object with no markdown fences and no extra text:
{
  "should_evolve": <true_or_false>,
  "analysis": "<short but concrete explanation>",
  "design_requirement": "<empty string if should_evolve is false, otherwise a concrete design requirement for the next guide/tool generation step>"
}
\end{lstlisting}

\begin{lstlisting}[style=promptstyle,caption={Compositional tool-reuse selection prompt.}]
{{AUTHORING_COMMON}}

You are now in an intermediate step before final guide/tool generation.

Do not write guide JSON or tool code yet. Your job is to identify up to 3 existing tools that are the best reuse candidates for the current design requirement.

Current design requirement:
{{DESIGN_REQUIREMENT}}

Existing available tools (TOOL_NAME + TOOL_SPEC):
{{EXISTING_TOOL_DEFINITIONS}}

Selection rules:
- choose at most 3 existing tools
- only choose from the provided `TOOL_NAME` values
- prefer tools whose `TOOL_SPEC.function.description` and parameter schema indicate they can directly support the requirement
- prefer a small, focused candidate set rather than a broad noisy list
- Do not reuse tools just for the sake of reuse; if the requirement calls for a meaningfully different workflow, returning an empty list is acceptable and may leave more room for new tool design.
- Do not choose multiple near-duplicate tools together, especially multiple similar reader tools or multiple similar writer tools; such near-duplicates usually should not appear together in the candidate set.
- Prefer complementary tools that can support a multi-stage workflow rather than a homogeneous bundle of tools with nearly identical roles.
- If the requirement suggests intermediate files, staged state transitions, or a new split of tool responsibilities, only reuse tools that clearly fit one stage of that design; leave the remaining stages open for new tools.
- In your analysis, pay attention to whether a tool can support steps such as reading raw input, writing intermediate state, rereading intermediate state, merging intermediate outputs, validating output, or writing the final result.
- if no existing tool is a good fit, return an empty list
- do not invent tool names
- do not return source code
- do not return guide JSON

Return exactly one JSON object with no markdown fences and no extra text:
{
  "candidate_tool_names": ["<TOOL_NAME_1>", "<TOOL_NAME_2>"],
  "analysis": "<short explanation of why these tools are the best reuse candidates, or why none fit>"
}
\end{lstlisting}

\begin{lstlisting}[style=promptstyle,caption={MemGuide/MemTool authoring prompt.}]
{{AUTHORING_COMMON}}

You are now in the final generation step.

Current design requirement:
{{DESIGN_REQUIREMENT}}

Candidate reusable existing tools (TOOL_NAME + TOOL_SPEC):
{{REUSABLE_TOOL_DEFINITIONS}}

Candidate reusable existing tool source code:
{{REUSABLE_TOOL_SOURCE_CODE}}

Generation instructions:
- Prefer reusing the provided candidate existing tools whenever they can satisfy the design requirement.
- If an existing tool is sufficient, reference its existing `TOOL_NAME` in the returned MemGuide and do not recreate that tool.
- Only create new MemTool code when the provided existing tools are not enough, are incompatible with the requirement, or have concrete shortcomings.
- It is valid to return only one MemGuide JSON block if no new MemTool code is needed.
- It is also valid to return one MemGuide JSON block plus one or more new MemTool Python code blocks if new tools are required.
- If you decide not to reuse a provided existing tool, make that decision implicitly through the returned workflow. Do not output explanations outside the required code blocks.

Required output format:
- Return only the required content, and nothing else. Do not include explanations, design notes, filename comments, or extra prose.
- You must first return exactly one MemGuide JSON block.
- If the current task requires MemTool code, append one or more complete Python code blocks after the MemGuide JSON block.
- If the current task does not require MemTool code, return only the MemGuide JSON block.
- The first non-whitespace characters of your answer must be exactly ```json
- Do not return raw JSON without a fenced code block.
- Do not wrap the whole answer in any outer markdown section, list item, heading, or explanation.
- If you need to return MemTool code, each MemTool must appear in its own separate ```python fenced code block after the JSON block.
- If your answer is not formatted as fenced code blocks exactly as required below, it is invalid.
\end{lstlisting}

\begin{lstlisting}[style=promptstyle,caption={MemGuide category-classification prompt.}]
You are classifying one newly created MemGuide into an existing set of guide categories.

Each existing category is represented by one representative guide. For each representative guide, you are given:
- `guide_file`
- `guide_name`
- `prompt`
- `tool_names`
- each referenced tool's `TOOL_NAME` and `TOOL_SPEC`

You are also given one new guide candidate with the same fields.

Existing guide category representatives:
{{GUIDE_CATEGORY_REPRESENTATIVES}}

New guide candidate:
{{NEW_GUIDE_CANDIDATE}}

Classification task:
- Decide whether the new guide candidate is highly similar to one existing representative guide category.
- Focus on workflow intent, processing strategy, guide prompt behavior, and the role of the referenced tools.
- Do not focus on filenames or small wording differences alone.
- If two guides share the same high-level state machine, for example both are still "read raw input -> LLM processes -> write final output" two-stage workflows, they should usually be classified into the same category even if tool names, field formatting, headings, or prompt wording differ.
- If the new guide only renames existing reader or writer tools, changes return formatting, adds a few extra fields, or lightly rewrites the prompt while keeping the same stage structure and tool responsibilities, it should still be treated as the existing category.
- Only classify a guide as a new category when it introduces substantial change in stage count, state transitions, intermediate-file usage, tool-type composition, or responsibility boundaries across tools.
- Pay special attention to whether it introduces explicit intermediate artifacts, longer tool chains, or a genuinely different division of roles among tools.
- If the new guide belongs to an existing category, return that representative guide's `guide_file`.
- If the new guide is meaningfully different from every existing category, return `null` so it can become a new category representative.

Required output:
- Return exactly one JSON object and nothing else.
- The JSON object must contain:
  - `matched_representative_guide_file`
  - `analysis`
- `matched_representative_guide_file` must be either one of the provided representative `guide_file` values or `null`.
- `analysis` must be a short explanation.

Valid output example:
```json
{
  "matched_representative_guide_file": "guide_3.json",
  "analysis": "The new guide follows the same extract-then-write workflow pattern and uses tools with the same functional role."
}
```

If no existing category matches, return:
```json
{
  "matched_representative_guide_file": null,
  "analysis": "The new guide introduces a distinct workflow pattern and should become a new category representative."
}
```
\end{lstlisting}

\begin{lstlisting}[style=promptstyle,caption={Diversity-guided exploration prompt.}]
{{AUTHORING_COMMON}}

You are now in the exploration generation step.

Current guide category representatives:
{{GUIDE_CATEGORY_REPRESENTATIVES}}

Exploration objective:
- Create one new MemGuide, and optionally one or more new MemTools, for the data-process subagent.
- The new guide must define a workflow pattern that is clearly different from every existing guide category representative above.
- Do not make only superficial edits such as renaming tools or lightly rewording the prompt.
- The difference should come from workflow design, decomposition strategy, reasoning structure, or tool responsibilities.
- By default, interpret "clearly different" to mean different stage count, different state transitions, different intermediate artifacts, different tool-role decomposition, or a meaningfully different information flow through the workflow.
- If the new design is still just "read raw input -> LLM directly processes -> write final output", it is usually not different enough unless it also introduces a genuinely different tool-role structure or intermediate-state mechanism.
- Prefer exploring multi-stage workflows over shortest-path workflows; when appropriate, explicitly design 3 or more stages.
- Prefer exploring combinations of different tool types such as extraction tools, structured-reorganization tools, intermediate-file writers, intermediate-file readers, fragment-merging tools, validation tools, and final-output writers.
- Strongly favor saving important intermediate results inside `item_dir` and having later tools or the LLM reread them to create clearer stage boundaries.
- Do not fake a new workflow by cloning existing `read_*` / `write_*` tools and changing only small wording details, field headers, or return strings.
- If existing tools are reused, reuse should usually be only a small part of the new workflow; the main novelty should come from new stage structure, tool responsibilities, and intermediate-state design.
- Reusing existing tool names inside the returned MemGuide is allowed only when that truly supports a clearly different guide category.
- If a genuinely new workflow requires new tools, create them.

Examples of encouraged exploration directions:
- "read raw input -> extract into intermediate JSON/TXT -> reread intermediate state and produce a compressed version -> reread the compressed version and write final output"
- "split original content -> summarize chunks -> save merged notes -> reread notes to generate the final result"
- "extract evidence and metadata first -> save candidate conclusions or checklists -> reread that checklist for validation/rewrite -> write final output"
- "separate read / extract / save-intermediate-state / reread-intermediate-state / finalize-write responsibilities across different tools instead of pushing all preprocessing into one reader tool"

Required output format:
- Return only the required content, and nothing else. Do not include explanations, design notes, filename comments, or extra prose.
- You must first return exactly one MemGuide JSON block.
- If the current task requires MemTool code, append one or more complete Python code blocks after the MemGuide JSON block.
- If the current task does not require MemTool code, return only the MemGuide JSON block.
- The first non-whitespace characters of your answer must be exactly ```json
- Do not return raw JSON without a fenced code block.
- Do not wrap the whole answer in any outer markdown section, list item, heading, or explanation.
- If you need to return MemTool code, each MemTool must appear in its own separate ```python fenced code block after the JSON block.
- If your answer is not formatted as fenced code blocks exactly as required below, it is invalid.
\end{lstlisting}

\end{document}